\documentclass{article}

\newif\ifarxiv
\arxivtrue

\ifarxiv
   \usepackage[accepted]{icml2020-no-icml-footer}
\else
   \usepackage[accepted]{icml2020}
\fi

\usepackage[utf8]{inputenc}   
\usepackage[T1]{fontenc}      
\usepackage{xcolor}           
\usepackage[
   colorlinks,
   citecolor=teal]{hyperref}  
\usepackage{url}              
\usepackage{booktabs}         
\usepackage{amsfonts}         
\usepackage{nicefrac}         
\usepackage{microtype}        
\usepackage{graphicx}         
\usepackage{hhline}           
\usepackage{makecell}         
\usepackage{multirow}         
\usepackage{comment}          
\usepackage{enumitem}	      
\usepackage[percent]{overpic} 
\usepackage{hyperref}
\usepackage{url}
\usepackage{times}
\usepackage{amssymb}
\usepackage{flushend}
\usepackage{xspace}

\usepackage{xspace}
\usepackage[textsize=footnotesize,textwidth=5em]{todonotes}

\usepackage{amsmath,amsfonts,bm}

\def\eqref#1{equation~\ref{#1}}

\def\1{\bm{1}}

\DeclareMathAlphabet{\mathsfit}{\encodingdefault}{\sfdefault}{m}{sl}
\SetMathAlphabet{\mathsfit}{bold}{\encodingdefault}{\sfdefault}{bx}{n}

\def\eg{\emph{e.g.,}}           
\def\ie{\emph{i.e.,}}           
\newcommand{\HLINE}{\Xhline{4\arrayrulewidth}}       
\newcommand{\tstrut}{\rule{0pt}{2.0ex}}              

\title{Multigrid Neural Memory}

\begin{document}

\twocolumn[
\icmltitle{Multigrid Neural Memory}



\icmlsetsymbol{equal}{*}

\begin{icmlauthorlist}
\icmlauthor{Tri Huynh}{uc}
\icmlauthor{Michael Maire}{uc}
\icmlauthor{Matthew R.\ Walter}{ttic}
\end{icmlauthorlist}

\icmlaffiliation{uc}{University of Chicago, Chicago, IL, USA}
\icmlaffiliation{ttic}{Toyota Technological Institute at Chicago, Chicago, IL, USA}

\icmlcorrespondingauthor{Tri Huynh}{trihuynh@uchicago.edu}


\vskip 0.3in
]



\printAffiliationsAndNotice{}  

\begin{abstract}
We introduce a novel approach to endowing neural networks with emergent, long-term, large-scale memory. Distinct from strategies that connect neural networks to external memory banks via intricately crafted controllers and hand-designed attentional mechanisms, our memory is internal, distributed, co-located alongside computation, and implicitly addressed, while being drastically simpler than prior efforts. Architecting networks with multigrid structure and connectivity, while distributing memory cells alongside computation throughout this topology, we observe the emergence of coherent memory subsystems. Our hierarchical spatial organization, parameterized convolutionally, permits efficient instantiation of large-capacity memories, while multigrid topology provides short internal routing pathways, allowing convolutional networks to efficiently approximate the behavior of fully connected networks.  Such networks have an implicit capacity for internal attention; augmented with memory, they learn to read and write specific memory locations in a dynamic data-dependent manner. We demonstrate these capabilities on exploration and mapping tasks, where our network is able to self-organize and retain long-term memory for trajectories of thousands of time steps. On tasks decoupled from any notion of spatial geometry: sorting, associative recall, and question answering, our design functions as a truly generic memory and yields excellent results.

\end{abstract}

\section{Introduction}
\label{sec:intro}

\vspace{-5pt}
Memory, in the form of generic, high-capacity, long-term storage, is likely to
play a critical role in expanding neural networks to new application domains.
A neural memory subsystem with such properties could be transformative---
pushing neural networks within grasp of tasks traditionally
associated with general intelligence and an extended sequence of reasoning
steps.  Development of architectures for integrating memory units with neural
networks spans a good portion of the history of neural networks themselves
(\eg~from LSTMs~\citep{LSTM} to the recent Neural Turing Machines
(NTMs)~\citep{NTM}).  Yet, while useful, none has elevated neural networks to
be capable of learning from and processing data on size and time scales
commensurate with traditional computing systems.  Recent successes of deep
neural networks, though dramatic, are focused on tasks, such as visual
perception or natural language translation, with relatively short latency---
\eg~hundreds of steps, often the depth of the network itself.

We present a network architecture that allows memory subsystems to emerge as a
byproduct of training simple components.  Though our networks appear
structurally uniform, they learn to behave like coherent large-scale memories,
internally coordinating a strategy for directing reads and writes to specific
layers and spatial locations therein.  As an analogy, a convolutional neural
network (CNN), tasked with classifying images, may coordinate and specialize
its internal layers for extracting useful visual representations.  Our
networks, trained for a task requiring long-term memory, do the same with
respect to memory: they self-organize their internal layers into a large-scale
memory store.  We accomplish this using only generic components; our networks
are comprised of LSTM cells and convolutional operations.  Yet, our networks
learn to master tasks that are beyond the abilities of traditional CNNs or
LSTMs.

\begin{figure*}[t]
   \begin{center}
      \includegraphics[width=1.0\linewidth]{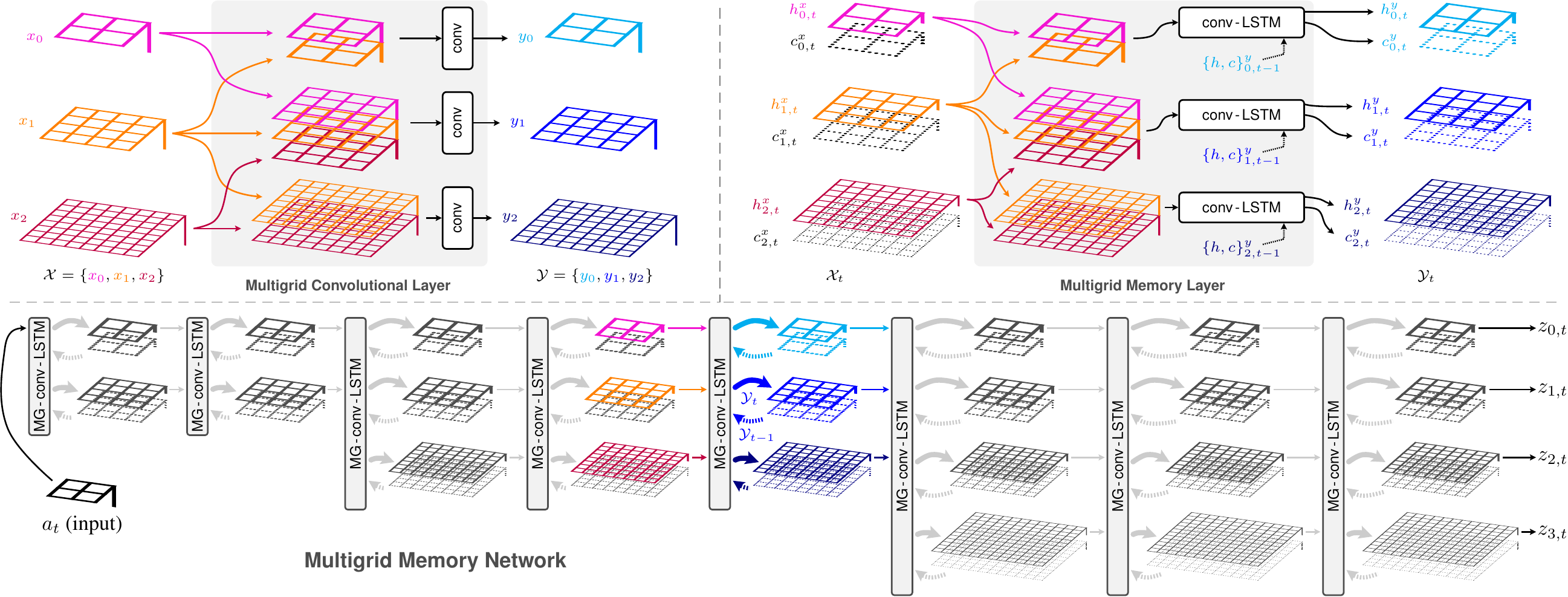}
   \end{center}
   \vspace{-10pt}
   \caption{
      \textbf{Multigrid memory architecture.}
      \emph{\textbf{Top Left:}}
         A multigrid convolutional layer~\citep{nmg} transforms input pyramid
         $\mathcal{X}$, containing activation tensors $\{x_0,x_1,x_2\}$, into
         output pyramid $\mathcal{Y}$ via learned filter sets that act across
         the concatenated representations of neighboring spatial scales.
      \emph{\textbf{Top Right:}}
         We design an analogous variant of the convolutional LSTM~\citep{
         convLSTM}, in which $\mathcal{X}$ and $\mathcal{Y}$ are indexed by
         time and encapsulate LSTM internals, \ie~memory cells ($c$) and hidden
         states ($h$).
      \emph{\textbf{Bottom:}}
         Connecting many such layers, both in sequence and across time, yields
         a multigrid mesh capable of routing input $a_t$ into a much larger
         memory space, updating a distributed memory representation, and
         providing multiple read-out pathways (\ie~$z_{0,t}$, $z_{1,t}$,
         $z_{2,t}$, $z_{3,t}$, or any combination thereof).
   }
   \label{fig:arch_mem}
   \vspace{-3pt}
\end{figure*}

Multigrid organization, imposed on both spatial layout and connectivity, is
the design principle that endows networks with this qualitatively new capacity
for forming self-organized memory subsystems.  Compared to almost all existing
networks, a multigrid wiring pattern provides an exponentially more efficient
routing topology among local components embedded within it.  \citet{nmg}
implement a multigrid variant of CNNs, demonstrating that efficient routing
capacity enables the network to learn tasks that require attentional behavior.
We distribute memory cells throughout such a network, and observe that this
implicit capacity for attention translates into an implicit capacity for
attention over memory read and write locations.  Learned parameters govern how
information flows through the network, what memory cells to update, and how to
update them.

Our design philosophy starkly contrasts with recent neural memory
architectures, including NTMs and the subsequent Differentiable Neural Computer
(DNC)~\citep{DNC}.  These prior approaches isolate memory in an external
storage bank, accessed via explicit addressing modes driven by custom
hand-crafted controllers; they graft a von Neumann memory model onto a neural
network.  Instead, we intertwine memory units throughout the interior of a deep
network.  Memory is a first-class citizen, rather than a separate data store
accessed via a special controller.  We introduce a new kind of network
layer---a multigrid memory layer---and use it as a stackable building block to
create deep memory networks.  Contrasting with simpler LSTMs, our memory is
truly deep; accessing an arbitrary memory location requires passing through
several layers.  Figure~\ref{fig:arch_mem} provides a visualization; we defer
the full details to Section~\ref{sec:method}.

There are major benefits to our design strategy, in particular:

\vspace{-10.0pt}
\begin{itemize}[leftmargin=0.2in]
   \item{
      \textbf{\emph{Distributed, co-located memory and compute.}}
      Our memory layers incorporate convolutional and LSTM components.
      Stacking such layers, we create not only a memory network, but also a
      generalization of both CNNs and LSTMs.  Our memory networks are
      standard networks with additional capabilities.
      Section~\ref{sec:experiments} shows they can learn tasks that require
      performing classification alongside storage and recall.

      This unification also opens a design space for connecting our memory
      networks to each other, as well as to standard networks.  Within a larger
      system, we can easily plug the internal state of our memory into a
      standard CNN---essentially granting that CNN read-only memory access.
      Sections~\ref{sec:method} and~\ref{sec:experiments} develop and
      experimentally validate two such memory interface approaches.
   }
   \vspace{-5.0pt}
   \item{
      \textbf{\emph{Scalability.}}
      Distributing storage over a multigrid hierarchy allows us to instantiate
      large amounts of memory while remaining parameter-efficient.  The
      low-level mechanism underlying memory access is convolution, and we
      inherit the parameter-sharing efficiencies of CNNs.  Our filters act
      across a spatially organized collection of memory cells, rather than the
      spatial extent of an image.  Increasing feature channels per memory cell
      costs parameters, but adding more cells incurs no such cost, decoupling
      memory size from parameter count.  Connecting memory layers across
      spatial pyramid levels allows for growing memory spatial extent
      exponentially with network depth, while guaranteeing there is a pathway
      between the network input and every memory unit.
   }
   \vspace{-15.0pt}
   \item{
      \textbf{\emph{Simplicity: implicit addressing, emergent subsystems.}}
      Our networks, once trained, behave like memory subsystems---this is an
      emergent phenomenon.  Our design contains no explicit address calculation
      unit, no controller, and no attention mask computation.  We take well known
      building blocks (\ie~convolution and LSTMs), wrap them in a multigrid wiring
      pattern, and achieve capabilities superior to those of the DNC, a far
      more complex design.
   }
\end{itemize}

\vspace{-5.0pt}
A diverse array of synthetic tasks serves as our experimental testbed.  Mapping
and localization, an inherently spatial task with relevance to robotics, is
one focus.  However, we avoid only experimenting with tasks naturally fit to
the architecturally-induced biases of our memory networks.  We also train them
to perform algorithmic tasks, as well as natural language processing (NLP)
tasks, previously used in analyzing the capabilities of NTMs and DNCs.
Throughout all settings, DNC accuracy serves as a baseline.  We observe
significant advantages for multigrid memory, including:

\vspace{-10.0pt}
\begin{itemize}[leftmargin=0.2in]
   \item{
      \textbf{\emph{Long-term retention.}}
      On spatial mapping tasks, our network correctly remembers observations
      of an external environment collected over paths thousands of time steps
      long.  Visualizing internal memory unit activations reveals an
      interpretable representation and algorithmic strategy our network learns
      for solving the problem.  The DNC, in contrast, fails to master these
      tasks.
   }
   \vspace{-5.0pt}
   \item{
      \textbf{\emph{Generality.}}
      On tasks decoupled from any notion of spatial geometry, such as
      associative recall or sorting (algorithmic), or question answering (NLP),
      our memory networks prove equally or more capable than DNCs.
   }
\end{itemize}

\vspace{-5.0pt}
Section~\ref{sec:experiments} further elaborates on experimental results.
Section~\ref{sec:conclusion} discusses implications: multigrid connectivity is
a uniquely innovative design principle, as it allows qualitatively novel behaviors
(attention) and subsystems (memory stores) to emerge from training simple
components.

\section{Related Work}
\label{sec:related}

\vspace{-5pt}
An extensive history of work seeks to grant neural networks the ability to
read and write memory~\citep{das1992learning, das1993using,
mozer1993connectionist, zeng1994discrete, holldobler1997designing}.
\citet{das1992learning} propose a neural pushdown automaton, which performs
differential push and pop operations on external memory.
\citet{schmidhuber1992learning} uses two feedforward networks: one produces
context-dependent weights for the second, whose weights may change quickly and
can be used as a form of memory.  \citet{schmidhuber1993self} proposes memory
addressing in the form of a ``self-referential'' recurrent neural network that
modifies its own weights.

Recurrent Long Short-Term Memory networks (LSTMs) \citep{LSTM} have
enabled significant progress on a variety of sequential prediction tasks,
including machine translation~\citep{sutskever2014sequence}, speech
recognition~\citep{DeepRNN}, and image captioning~\citep{RCNN}.  LSTMs are
Turing-complete~\citep{siegelmann1995computational} and are, in principle,
capable of context-dependent storage and retrieval over long time
periods~\citep{DeepRNN:NIPS}.  However, capacity for long-term read-write is
sensitive to the training procedure~\citep{RNNCap} and is limited in practice.

Grid LSTMs~\citep{kalchbrenner2015grid} arrange LSTM cells in a 2D or 3D grid,
placing recurrent links along all axes of the grid.  This sense of grid
differs from our usage of multigrid, as the latter refers to links across a
multiscale spatial layout.  In~\citet{kalchbrenner2015grid}'s terminology, our
multigrid memory networks are not Grid LSTMs, but are a variant of Stacked
LSTMs~\citep{DeepRNN}.

To improve the long-term read-write abilities of recurrent networks, several
modifications have been proposed.  These include differentiable attention
mechanisms~\citep{graves2013generating, bahdanau2014neural, mnih2014recurrent,
vqa_sat} that provide a form of content-based memory addressing, pointer
networks~\citep{vinyals2015pointer} that ``point to'' rather than blend
inputs, and architectures that enforce independence among neurons within each
layer~\citep{IndRNN}.

A number of methods augment the short- and long-term memory internal to
recurrent networks with external ``working'' memory, in order to realize
differentiable programming architectures that can learn to model and execute
various programs~\citep{NTM, DNC, MemNet, sukhbaatar2015end, StackRNN,
reed2015neural, grefenstette2015learning, kurach2015neural}.  Unlike our
approach, these methods explicitly decouple memory from computation, mimicking
a standard computer architecture.  A neural controller (analogous to a CPU)
interfaces with specialized external memory (\eg~random-access memory or
tapes).

The Neural Turing Machine~\citep{NTM} augments neural networks with a hand-designed
attention mechanism to read from and write to external memory in a
differentiable fashion.  This enables the NTM to learn to perform various
algorithmic tasks, including copying, sorting, and associative recall.  The
Differential Neural Computer~\citep{DNC} improves upon the NTM with support for
dynamic memory allocation and additional memory addressing modes.  Without a
sparsifying approximation, DNC runtime grows quadratically with memory due to
the need to maintain the temporal link matrix.  Our architecture has no such
overhead, nor does it require maintaining any auxiliary state.

Other methods enhance recurrent layers with differentiable forms of a
restricted class of memory structures, including stacks, queues, and
dequeues~\citep{grefenstette2015learning,StackRNN}.
\citet{gemici2017generative} augment structured dynamic models for temporal
processes with various external memory architectures~\citep{NTM, DNC,
santoro2016one}.

Similar memory-explicit architectures have been proposed for deep reinforcement
learning (RL) tasks.  While deep RL has been applied to several challenging
domains~\citep{mnih2015human, hausknecht2015deep, levine2016end}, most
approaches reason over short-term state representations, which limits their
ability to deal with partial observability inherent in many tasks.  Several
methods augment deep RL architectures with external memory to facilitate
long-term reasoning.  \citet{oh2016control} maintain a fixed number of recent
states in memory and then read from the memory using a soft attention
operation.  \citet{NeuralMap} propose a specialized write operator, together
with a hand-designed 2D memory structure, both specifically crafted for
navigation in maze-like environments.

Rather than learn when to write to memory (\eg~as done by NTM and DNC),
\citet{EpControl} continuously write the experience of an RL agent to a
dictionary-like memory module queried in a key-based fashion (permitting large
memories).  Building on this framework, \citet{GenModelMem} augment a
generative temporal model with a specialized form of spatial memory that
exploits privileged information, including an explicit representation of the
agent's position.

Though we experiment with RL, our memory implementation contrasts with
this past work.  Our multigrid memory architecture jointly couples
computation with memory read and write operations, and learns how to use a
generic memory structure rather than one specialized to a particular task.

\section{Multigrid Memory Architectures}
\label{sec:method}


\vspace{-5pt}
A common approach to endowing neural networks with long-term memory builds
memory addressing upon explicit attention mechanisms.  Such attention
mechanisms, independent of memory, are hugely influential in natural language
processing~\citep{vaswani2017attention}.  NTMs~\citep{NTM} and DNCs~\citep{DNC}
address memory by explicitly computing a soft attention mask over memory
locations.  This leads to a design reliant on an external memory controller,
which produces and then applies that mask when reading from or writing to a
separate memory bank.

We craft a memory network without such strict division into modules. Instead,
we propose a structurally uniform architecture that generalizes modern
convolutional and recurrent designs by embedding memory cells within the
feed-forward computational flow of a deep network.  Convolutional neural
networks and LSTMs (specifically, the convolutional LSTM
variety~\citep{convLSTM}) exist as strict subsets of the full connection set
comprising our multigrid memory network.  We even encapsulate modern residual
networks~\citep{he2015deep}.  Though omitted from diagrams
(\eg~Figure~\ref{fig:arch_mem}) for the sake of clarity, we utilize residual
connections linking the inputs of subsequent layers across the depth (not time)
dimension of our memory networks.

In our design, memory addressing is implicit rather than explicit.  We build
upon an implicit capacity for attentional behavior inherent in a specific
kind of network architecture.  \citet{nmg} propose a multigrid variant of both
standard CNNs and residual networks (ResNets).  While their primary experiments
concern image classification, they also present a striking result on a
synthetic image-to-image transformation task: multigrid CNNs (and multigrid
ResNets) are capable of learning to emulate attentional behavior.  Their
analysis reveals that the multigrid connection structure is both essential to
and sufficient for enabling this phenomenon.

The underlying cause is that bi-directional connections across a scale-space
hierarchy (Figure~\ref{fig:arch_mem}, left) create exponentially shorter
signalling pathways between units at different locations on the spatial grid.
Specifically, coarse-to-fine and fine-to-coarse connections between pyramids in
subsequent layers allow a signal to hop up pyramid levels and back down again
(and vice-versa).  As a consequence, pathways connect any neuron in a given
layer with every neuron located only $\mathcal{O}(\log(S))$ layers deeper,
where $S$ is the spatial extent (diameter) of the highest-resolution grid (see
\ifarxiv
Appendix~\ref{sec:app_information_routing}
\else
supplementary material
\fi
 for a detailed analysis).  In a standard
convolutional network, this takes $\mathcal{O}(S)$ layers.  These shorter
pathways enable our convolutional architecture to approximate the behavior of
a fully-connected network.

%


By replacing convolutional layers with convolutional LSTMs~\citep{convLSTM},
we convert the inherent attentional capacity of multigrid CNNs into an inherent
capacity for distributed memory addressing.  Grid levels no longer correspond
to operations on a multiresolution image representation, but instead correspond
to accessing smaller or larger storage banks within a distributed memory
hierarchy.  Dynamic routing across scale space (in the multigrid CNN)
now corresponds to dynamic routing into different regions of memory, according
to a learned strategy.  

\vspace{-5pt}
\subsection{Multigrid Memory Layer}
\label{sec:mem_layer}

\vspace{-5pt}
Figure~\ref{fig:arch_mem} diagrams both the multigrid convolutional layer of
\citet{nmg} and our corresponding multigrid memory (MG-conv-LSTM) layer.
Activations at a particular depth in our network consist of a pyramid
$\mathcal{X}_t = \{(h^{x}_{j,t},c^{x}_{j,t})\}$,
where $j$ indexes the pyramid level, $t$ indexes time, and $x$ is the layer.  $h^x$ and  $c^x$
denote the hidden state and memory cell contents of a convolutional
LSTM~\citep{convLSTM}, respectively.  Following the construction of~\citet{nmg},
states $h^x$ at neighboring scales are resized and concatenated, with the
resulting tensors fed as inputs to the corresponding scale-specific
convolutional LSTM units in the next multigrid layer.  The state associated
with a conv-LSTM unit at a particular layer and level
$(h^{y}_{j,t},c^{y}_{j,t})$
is computed from memory:
$h^{y}_{j,t-1}$ and $c^{y}_{j,t-1}$,
and input tensor:
${\uparrow}h^{x}_{j-1,t} \oplus h^{x}_{j,t} \oplus {\downarrow}h^{x}_{j+1,t}$,
where $\uparrow$, $\downarrow$, and $\oplus$ denote upsampling, downsampling,
and concatenation.  Specifically, a multigrid memory layer
(Figure~\ref{fig:arch_mem}, top right) operates as:

\vspace{-20pt}
\begin{small}
\begin{align*}
   H^{x}_{j,t} &:=
        ({\uparrow}h^{x}_{j-1,t}) \oplus
                    (h^{x}_{j,t}) \oplus
      ({\downarrow}h^{x}_{j+1,t}) \\
   i_{j,t} &:=
      \sigma(
         W^{xi}_{j} * H^{x}_{j,t} +
         W^{hi}_{j} * h^{y}_{j,t-1} +
         W^{ci}_{j} \circ c^{y}_{j,t-1} + b^i_j) \\
   f_{j,t} &:=
      \sigma(
         W^{xf}_{j} * H^{x}_{j,t} +
         W^{hf}_{j} * h^{y}_{j,t-1} +
         W^{cf}_{j} \circ c^{y}_{j,t-1} + b^f_j) \\
   c^{y}_{j,t} &:=
         f_{j,t} \circ c^{y}_{j,t-1} \!+
         i_{j,t} \circ
         \tanh(W^{xc}_{j} * H^{x}_{j,t} \!+\! W^{hc}_{j} \!* h^{y}_{j,t-1} \!+ b^c_j) \\
   o^{y}_{j,t} &:=
      \sigma(
         W^{xo}_{j} * H^{x}_{j,t} +
         W^{ho}_{j} * h^{y}_{j,t-1} +
         W^{co}_{j} \circ c^{y}_{j,t} + b^o_j) \\
   h^{y}_{j,t} &:=
      o^{y}_{j,t} \circ \tanh(c^{y}_{j,t})
\end{align*}
\end{small}

\vspace{-20pt}
Superscripts denote variable roles (\eg~layer $x$ or $y$, and/or a
particular parameter subtype for weights or biases).  Subscripts index pyramid
level $j$ and time $t$, $*$ denotes convolution, and $\circ$ the Hadamard product.
Computation resembles~\citet{convLSTM}, with additional input tensor assembly,
and repetition over output pyramid levels $j$.  If a particular input pyramid
level is not present in the architecture, it is dropped from the concatenation
in the first step.  Like \citet{nmg}, downsampling ($\downarrow$) includes
max-pooling.  We utilize a two-dimensional memory geometry, and change
resolution by a factor of two in each spatial dimension when moving up or down
a pyramid level.

Connecting many such memory layers yields a memory network or distributed
memory mesh, as shown in the bottom diagram of Figure~\ref{fig:arch_mem}.  Note
that a single time increment (from $t-1$ to $t$) involves running an entire
forward pass of the network, propagating the input signal $a_t$ to the deepest
layer $z_t$.  Though not drawn here, we also incorporate batch normalization
layers and residual connections along grids of corresponding resolution
(\ie~from $h^{x}_{j,t}$ to $h^{y}_{j,t}$).  These details mirror~\citet{nmg}.
The convolutional nature of the multigrid memory architecture, together with
its routing capability provides parameter-efficient implicit addressing of a
scalable memory space.

\vspace{-5pt}
\subsection{Memory Interfaces}
\label{sec:interfaces}

\begin{figure}[t]
   \begin{center}
      \begin{minipage}[t]{1.0\linewidth}
         \vspace{0pt}
         \includegraphics[angle=90,width=1.0\linewidth]{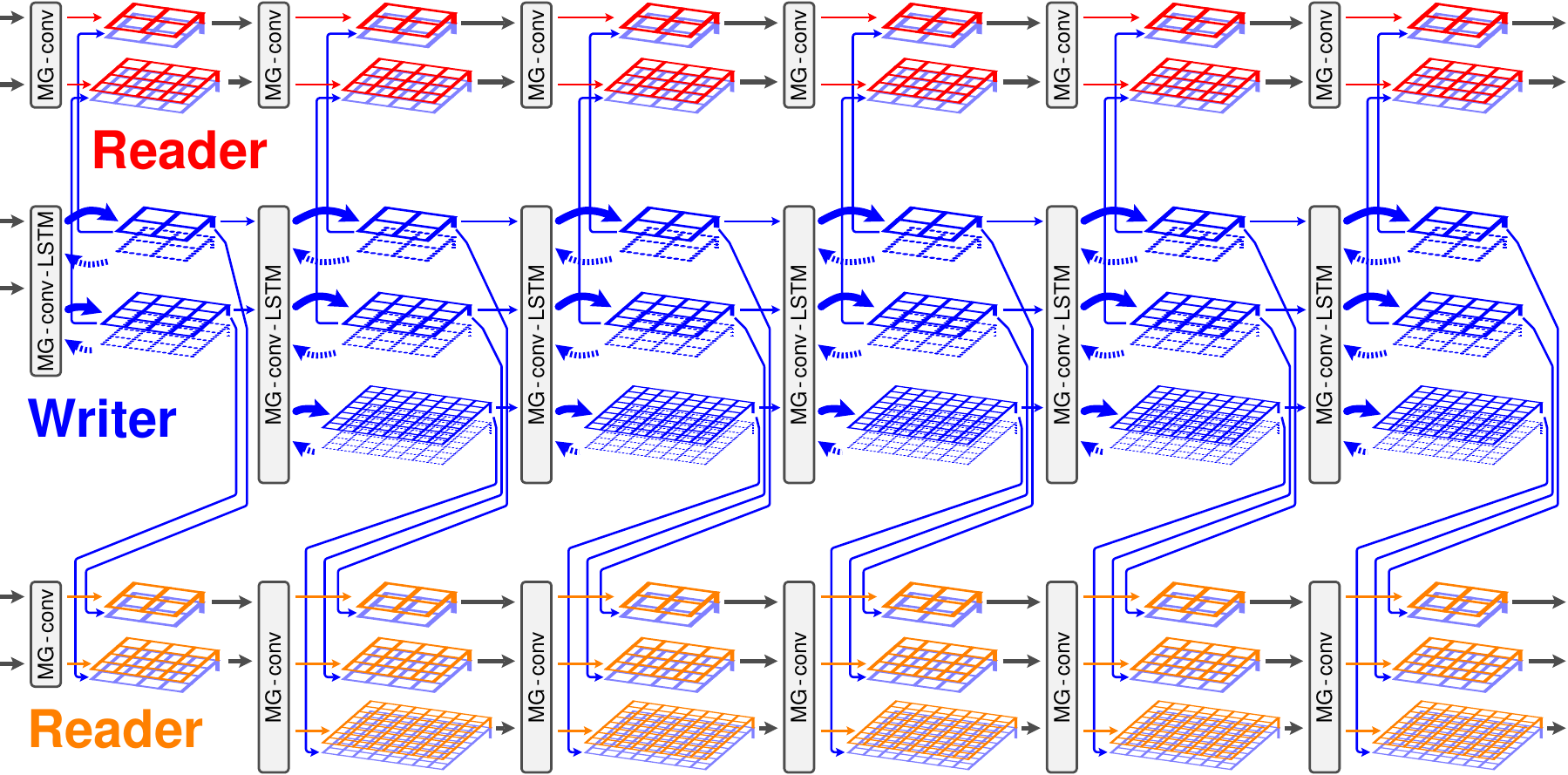}
      \end{minipage}
      \vspace{5pt}
      \hrule{}
      \vspace{5pt}
      \begin{minipage}[t]{1.0\linewidth}
         \includegraphics[angle=90,width=1.0\linewidth]{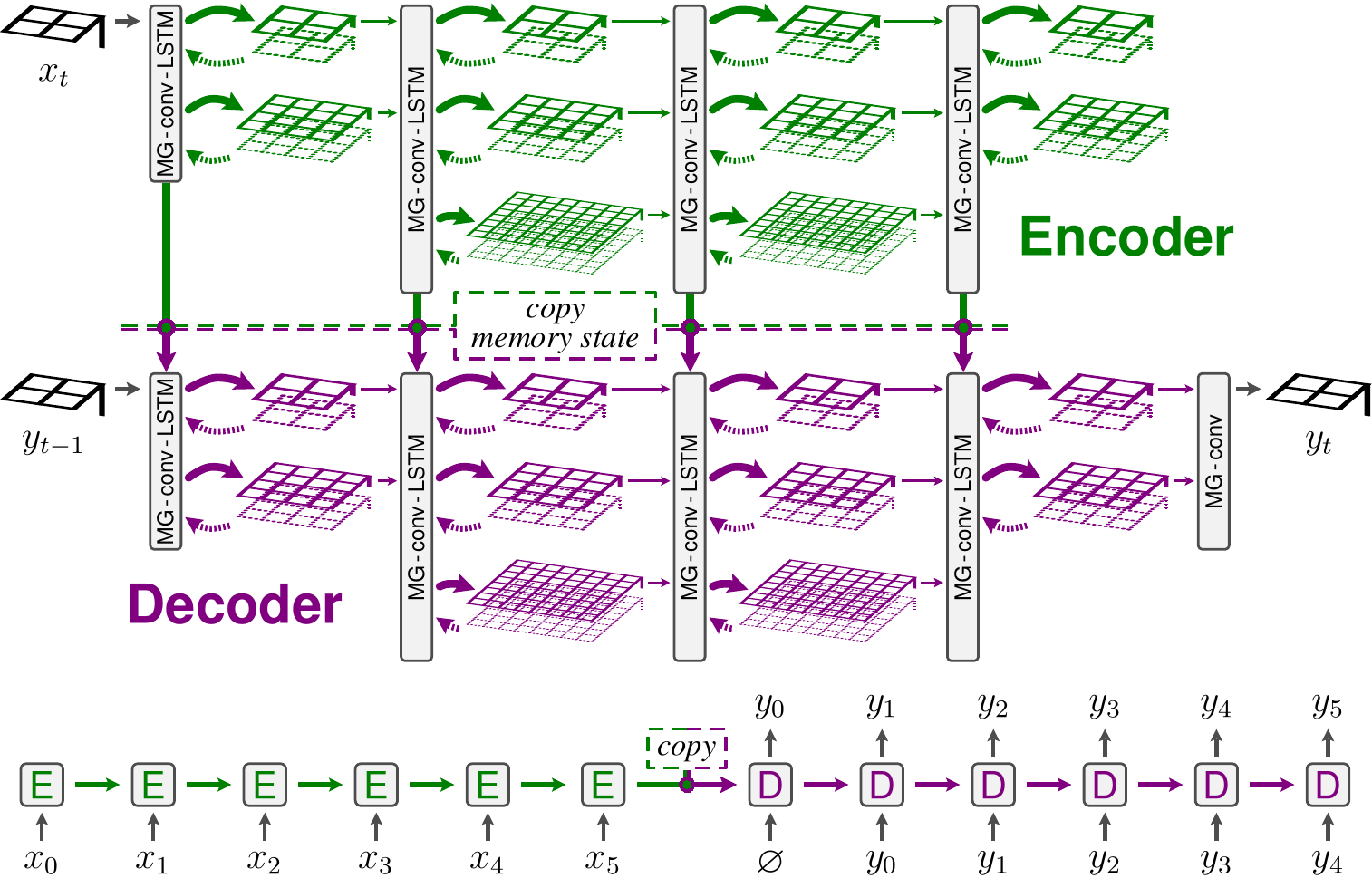}
      \end{minipage}
   \end{center}
   \vspace{-10pt}
   \definecolor{darkgreen}{rgb}{0,0.5,0}
   \caption{
      \textbf{Memory interfaces.}
      \emph{\textbf{Top:}}
         Multiple readers (\textcolor{red}{red}, \textcolor{orange}{orange})
         and a single \textcolor{blue}{writer} simultaneously manipulate
         a multigrid memory.  Readers are multigrid CNNs; each convolutional
         layer views the hidden state of the corresponding grid in memory by
         concatenating it as an additional input.
      \emph{\textbf{Bottom:}}
         Distinct \textcolor{darkgreen}{encoder} and
         \textcolor{violet}{decoder} networks, each structured as a deep
         multigrid memory mesh, cooperate to perform a sequence-to-sequence
         task.  We initialize the memory pyramid (LSTM internals) of each
         decoder layer by copying it from the corresponding encoder layer.
   }
   \label{fig:arch_nets}
   \vspace{-10pt}
\end{figure}

\vspace{-5pt}
As our multigrid memory networks are multigrid CNNs plus internal memory units,
we are able to connect them to other neural network modules as freely and
flexibly as one can do with CNNs.  Figure~\ref{fig:arch_nets} diagrams a few
such interface architectures, which we experimentally explore in
Section~\ref{sec:experiments}.

In Figure~\ref{fig:arch_nets} (top), multiple ``threads'', two readers and one
writer, simultaneously access a shared multigrid memory.  The memory itself is
located within the writer network (blue), which is structured as a deep
multigrid convolutional-LSTM.  The reader networks (red and orange), are merely
multigrid CNNs, containing no internal storage, but observing the hidden state
of the multigrid memory network.

Figure~\ref{fig:arch_nets} (bottom) diagrams a deep multigrid analogue
of a standard paired recurrent encoder and decoder.  This design substantially
expands the amount of memory that can be manipulated when learning
sequence-to-sequence tasks.

\section{Experiments}
\label{sec:experiments}

\begin{figure*}[t]
   \begin{center}
      \includegraphics[width=0.98\linewidth]{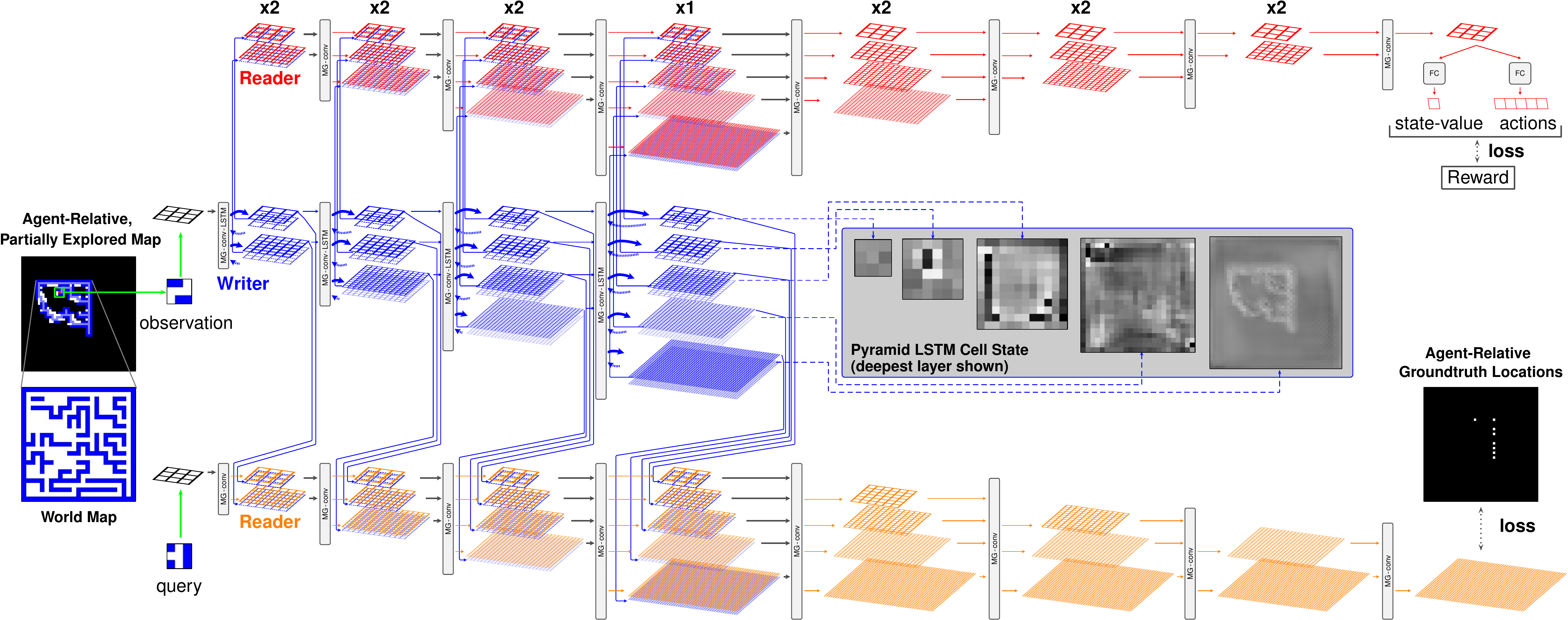}
   \end{center}
   \vspace{-10pt}
   \definecolor{darkgreen}{rgb}{0,0.5,0}
   \caption{
      \textbf{Mapping, localization, and exploration.}
      An agent is comprised of a deep multigrid \textcolor{blue}{memory}, and
      two deep multigrid CNNs (\textcolor{orange}{query} and
      \textcolor{red}{policy} subnetworks), which have memory read access.
      Navigating a maze, the agent makes a local observation at each time step,
      and chooses a next action, receiving reward for exploring unseen areas.
      Given a random local patch, the \textcolor{orange}{query} subnet must report
      all previously observed maze locations whose local observations match that patch.  Subnet
      colors reflect those in Figure~\ref{fig:arch_nets}.
   }
   \label{fig:maze}
\end{figure*}

\begin{figure}[t]
   \begin{center}
      \begin{overpic}[height=0.8in]{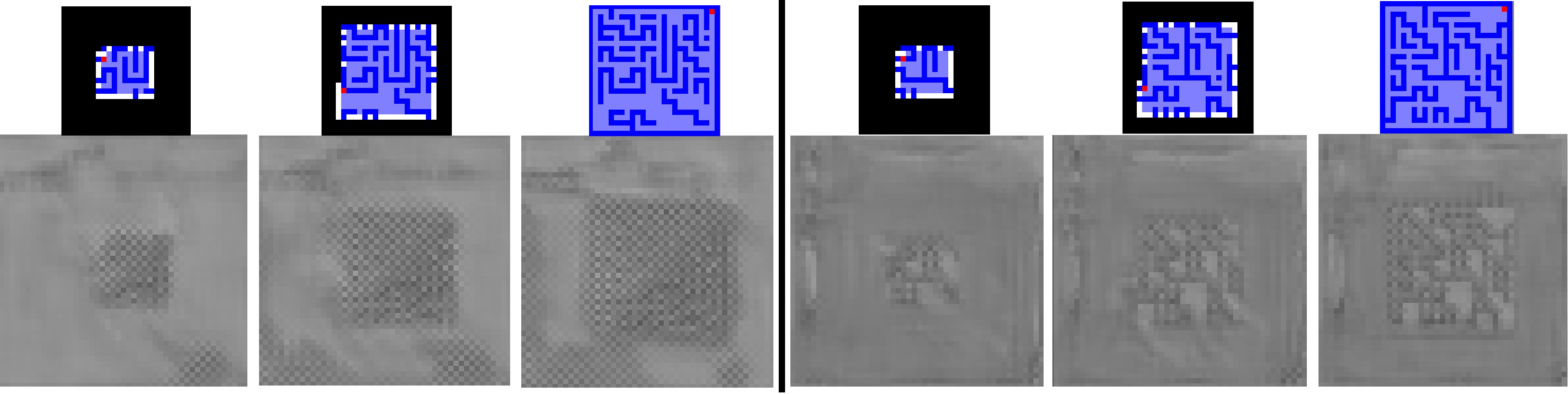}
         \put(5,-3){\tiny{\textsf{\textbf{Observation: 3$\times$3, Query: 3$\times$3}}}}
         \put(55,-3){\tiny{\textsf{\textbf{Observation: 3$\times$3, Query: 9$\times$9}}}}
      \end{overpic}
   \end{center}
   \vspace{-5pt}
   \definecolor{darkgreen}{rgb}{0,0.5,0}
   \caption{
      \textbf{Memory Visualization.}
      Memory contents (hidden states $\{h_{t}\}$ on deepest, highest-resolution grid) mirror the
      map explored with spiral motion (top), vividly showing the interpretable strategy for self-organizing, implicit attentional addressing (reading/writing) of highly specific memory cells when training localization tasks, without having hand-designed attention mechanisms.}
   \label{fig:learning_rep}
   \vspace{-10pt}
\end{figure}

\vspace{-5pt}
We evaluate our multigrid neural memory architecture on a diverse set of
domains.  We begin with a reinforcement learning-based navigation task, in
which memory provides a representation of the environment (\ie~a map).  To
demonstrate the generalizability of our memory architecture on domains decoupled from spatial geometry, we also consider
various algorithmic and NLP tasks previously used to evaluate the performance
of NTMs and DNCs.

\vspace{-5pt}
\subsection{Mapping \& Localization}
\label{sec:mapping}

We first consider a navigation problem, in which an agent explores a priori
unknown environments with access to only observations of its immediate
surroundings.  Effective navigation requires maintaining a consistent
representation of the environment (\ie~a map).  Using memory as a form of map,
an agent must learn where and when to perform write and read operations as it
moves, while retaining the map over long time periods.  This task mimics
partially observable spatial navigation scenarios considered by memory-based
deep reinforcement learning (RL) frameworks.

\emph{\textbf{Problem Setup:}} The agent navigates an unknown $n\times n$ 2D
maze and observes only the local $m \times m$ grid ($m \ll n$) centered at
the agent's position.  It has no knowledge of its absolute position.  Actions
consist of one-step motion in the four cardinal directions.  While
navigating, we query the network with a randomly chosen, previously seen
$k \times k$ patch ($m \le k \ll n$) and ask it to identify every location
matching that patch in the explored map.  See Figure~\ref{fig:maze}.

\emph{\textbf{Multigrid Architecture:}}
We use a deep multigrid network with multigrid memory and multigrid CNN
subcomponents (Figure~\ref{fig:maze}).  Our memory (writer subnet) consists of
seven MG-conv-LSTM layers, with pyramid spatial scales progressively increasing
from $3 \times 3$ to $48 \times 48$.  The reader, structured similarly, has an
output attached to its deepest $48 \times 48$ grid, and is tasked with answering
localization queries.  Section~\ref{sec:exploring} experiments with an
additional reader network that predicts actions, driving the agent to explore.

In order to understand the network's ability to maintain a ``map'' of the
environment in memory, we first consider a setting in which the agent executes
a pre-defined navigation policy, and evaluate its localization performance.  We
consider different policies (spiraling outwards and a random walk), different
patch sizes for observation and localization ($3\times3$ or $9\times9$), as well
as different trajectory (path) lengths.  We experiment in two regimes: small
($8$K) memory multigrid and DNC models, calibrated to the maximum trainable DNC
size, and larger memory multigrid and convolutional-LSTM variants.  We compare
against:
\begin{itemize}[leftmargin=0.2in,itemsep=2pt,topsep=0pt]
   \item{
      \textbf{Differentiable Neural Computer (DNC)}~\citep{DNC}:
      see details in
      \ifarxiv
         Appendix~\ref{sec:app_dnc}.
      \else
         the supplementary material.
      \fi
      }
   \vspace{-5pt}
   \item{
      \textbf{Ablated MG}: a multigrid architecture variant including only the
      finest pyramid scale at each layer.
   }
   \vspace{-5pt}
   \item{
      \textbf{ConvLSTM-deep}: a network made of $23$ convolutional-LSTM layers,
      each on a $48 \times 48$ grid, yielding the same total grid count as our
      $7$-layer multigrid network.
   }
   \vspace{-5pt}
   \item{
      \textbf{ConvLSTM-thick}: $7$ layers of convolutional-LSTMs acting on
      $48 \times 48$ grids.  We set channel counts to the sum of channels
      distributed across the corresponding pyramid layer of our large
      multigrid network.
   }
\end{itemize}

We train each architecture using RMSProp.  We search over learning rates in
log scale from $10^{-2}$ to $10^{-4}$, and use $10^{-3}$ for multigrid and
ConvLSTM, and $10^{-4}$ for DNC.  We use randomly generated maps  for training
and testing.  Training runs for $8\times10^6$ steps with batch size $32$.  Test
set size is $5000$ maps.  We used a pixel-wise cross-entropy loss over
predicted and true locations (see
\ifarxiv
Appendix~\ref{sec:app_spatial_map_loss}).
\else
supplementary material).
\fi

Table~\ref{tab:mapping} reports performance in terms of localization accuracy
on the test set.  For the $25 \times 25$ world in which the agent moves in a
spiral (\ie~predictable) motion and the observation and query are $3 \times 3$,
our small multigrid network achieves near perfect precision ($99.33\%$),
recall ($99.02\%$), and F-score ($99.17\%$), while all baselines struggle. Both
ConvLSTM baselines fail to learn the task; simply stacking convolutional-LSTM
units does not work.  DNC performs similarly to Ablated MG in terms of
precision ($\approx77.6\%$), at the expense of a significant loss in recall
($14.50\%$).  Instead, tasked with the simpler job of localization in a
$15 \times 15$ world, the performance of the DNC improves, yet its scores are still around
$10\%$ lower than those of multigrid on the more challenging $25 \times 25$
environment.  For the $25 \times 25$ map, efficiently addressing a large memory is
required; the DNC's explicit attention strategy appears to fall behind our
implicit routing mechanism.  Trying to compensate by augmenting the DNC with
more memory is difficult: without a sparsifying approximation, the DNC's
temporal memory linkage matrix incurs a quadratic cost in memory size (see
\ifarxiv
Appendix~\ref{sec:app_dnc}).
\else
supplementary material).
\fi
Our architecture has no such overhead, nor does it
require maintaining auxiliary state.  Even limiting multigrid to $8$K memory,
it has no issue mastering the $25 \times 25$ world.

Figure~\ref{fig:learning_rep} visualizes the contents of the deepest and
high-resolution LSTM block within the multigrid memory network of an agent
moving in a spiral pattern.  This memory clearly mirrors the contents of the
true map.  The network has learned a correct, and incidentally, an
interpretable, procedure for addressing and writing memory.

\begin{table*}[t]
   \begin{minipage}[t]{0.29\linewidth}
      \vspace{0pt}
      \caption{
         \textbf{Mapping and localization.}
         Our network significantly outperforms the DNC and other baselines.
         Efficient memory usage, enabled by multigrid connectivity, is
         essential; the DNC even fails to master smaller 15$\times$15 mazes.
         Our network retains memory over thousands of time-steps.  Our
         localization subnet, trained on random motion, generalizes to queries
         for a \emph{policy-driven agent} (last row).
      }
      \label{tab:mapping}
   \end{minipage}
   \hfill
   \begin{minipage}[t]{0.695\linewidth}
      \vspace{0pt}
      \begin{center}
         \begin{scriptsize}
         \setlength\tabcolsep{4.75pt}
         \begin{tabular}{@{}l|cc|c|ccc|c|ccc@{}}
\multirow{2}{*}{%
Architecture}               & Params            & Memory             &              World             &                          \multicolumn{3}{c|}{Task Definition}                         &           Path                      &                 \multicolumn{3}{c}{Localization Accuracy} \\
                            & $(\times10^6)$    & $(\times10^3)$     &              Map               &            FoV              &          Motion          &         Query                &          Length                     &           Prec.       &         Recall        &                   F \\
\HLINE\tstrut
MG Mem+CNN                  & 0.12              & \hphantom{00}7.99  &  \multirow{2}{*}{15$\times$15} & \multirow{8}{*}{3$\times$3} & \multirow{8}{*}{Spiral}  & \multirow{8}{*}{3$\times$3}  &\multirow{2}{*}{\hphantom{1}169}     &       \textbf{99.79}  &      \textbf{99.88}   &       \textbf{99.83} \\
DNC                         & 0.75              & \hphantom{00}8.00  &                                &                             &                          &                              &                                     &               91.09   &              87.67    &               89.35 \\
\cline{2-4}\cline{8-11}\tstrut
MG Mem+CNN                  & 0.17              & \hphantom{00}7.99  & \multirow{6}{*}{25$\times$25}  &                             &                          &                              & \multirow{6}{*}{\hphantom{1}529}    &       \textbf{99.33}  &      \textbf{99.02}   &       \textbf{99.17} \\
DNC                         & 0.68              & \hphantom{00}8.00  &                                &                             &                          &                              &                                     &               77.63   &              14.50    &               24.44 \\
\cline{2-3}\cline{9-11}\tstrut
MG Mem+CNN                  & 0.65              & \hphantom{0}76.97  &                                &                             &                          &                              &                                     &       \textbf{99.99}  &      \textbf{99.97}   &       \textbf{99.98} \\
Ablated MG                  & 1.40              & 265.54             &                                &                             &                          &                              &                                     &               77.57   &              51.27    &               61.73 \\
ConvLSTM-deep               & 0.38              & 626.69             &                                &                             &                          &                              &                                     &               43.42   &   \hphantom{0}3.52    &    \hphantom{0}6.51 \\
ConvLSTM-thick              & 1.40              & 626.69             &                                &                             &                          &                              &                                     &               47.68   &   \hphantom{0}1.11    &    \hphantom{0}2.16 \\
\HLINE\tstrut
\multirow{5}{*}{MG Mem+CNN} & 0.79              & \hphantom{0}76.97  & \multirow{5}{*}{25$\times$25}  & \multirow{1}{*}{3$\times$3} & \multirow{1}{*}{Spiral}  & \multirow{1}{*}{9$\times$9}  & \multirow{1}{*}{\hphantom{1}529}    &               97.34   &              99.50    &               98.41 \\
\cline{2-3}\cline{5-11}\tstrut
                            & \multirow{2}{*}{0.65}
                            &     \multirow{2}{*}{\hphantom{0}76.97} &                                & \multirow{2}{*}{3$\times$3} & \multirow{2}{*}{Random}  & \multirow{2}{*}{3$\times$3}  & \multirow{1}{*}{\hphantom{1}500}    &               96.83   &               95.59   &               96.20 \\
\cline{8-11}\tstrut
                            &                   &                    &                                &                             &                          &                              & \multirow{1}{*}{1500}               &               96.13   &               91.08   &               93.54 \\
\cline{2-3}\cline{5-11}\tstrut
                            & 0.66              & \hphantom{0}78.12  &                                & \multirow{1}{*}{9$\times$9} & \multirow{1}{*}{Random}  & \multirow{1}{*}{9$\times$9}  & \multirow{1}{*}{\hphantom{1}500}    &               92.82   &               87.60   &               90.14 \\
\cline{2-3}\cline{5-11}\tstrut
                            & 0.65              & \hphantom{0}76.97  &                                & \multirow{1}{*}{3$\times$3} & \textbf{\emph{Policy}}   & \multirow{1}{*}{3$\times$3}  & \multirow{1}{*}{1000}               & \textbf{\emph{95.65}} & \textbf{\emph{90.22}} & \textbf{\emph{92.86}} \\
\hline
         \end{tabular}
         \end{scriptsize}
      \end{center}
   \end{minipage}
\end{table*}

\begin{figure}[t]
  \centering
     \includegraphics[width=1.0\linewidth]{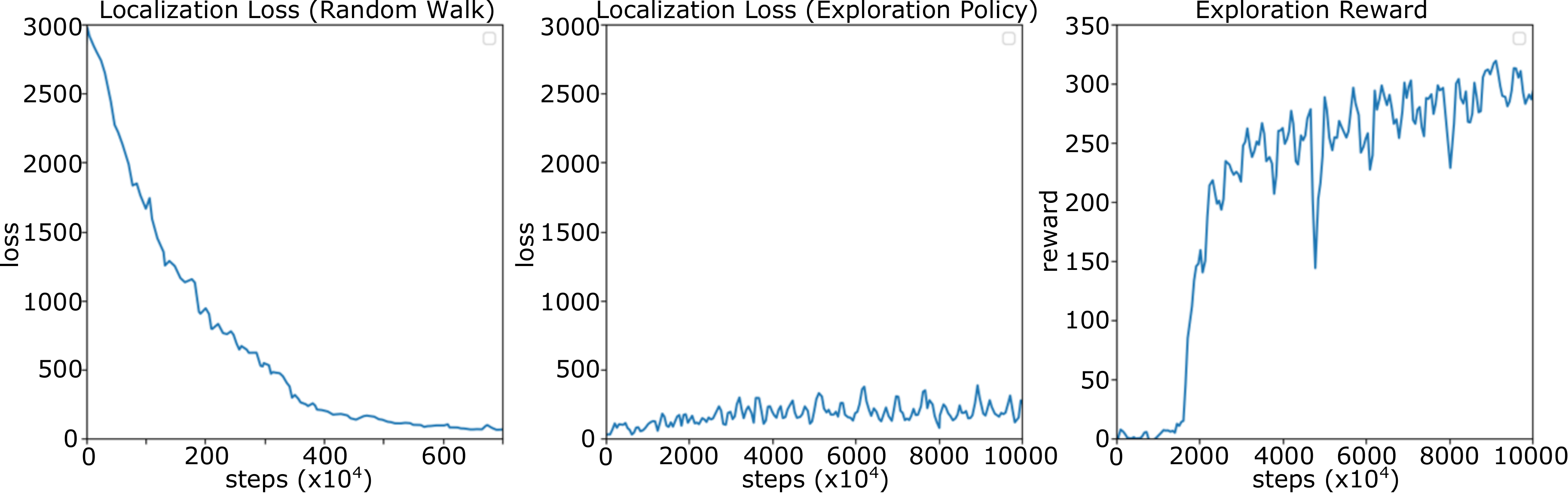}
     \vspace{-20pt}
     \caption{
        \textbf{Generalization of localization.}
        Fixing parameters after training the query subnet on random motion
        \emph{\textbf{(left)}}, its localization loss remains low while
        training the exploration policy \emph{\textbf{(middle)}}, whose reward
        improves \emph{\textbf{(right)}}.
     }
    \vspace{-15pt}
     \label{fig:generalization}
\end{figure}

In more complex settings for motion type and query size
(Table~\ref{tab:mapping}, bottom) our multigrid network remains accurate.  It
even generalizes to motions different from those on which it trained, including
motion dictated by the learned policy that we describe shortly.  Notably, even
with the very long trajectory of $1500$ time steps, our proposed architecture
has no issue retaining a large map memory.

\vspace{-5pt}
\subsection{Joint Exploration, Mapping, and Localization}
\label{sec:exploring}

\vspace{-5pt}
We now consider a setting in which the agent learns an exploration policy via
reinforcement, on top of a fixed mapping and localization network pre-trained
with random walk motion.  We implement the policy network as another multigrid
reader, and leverage the pre-trained mapping and localization capabilities to
learn a more effective policy.

We formulate exploration as a reinforcement learning problem: the agent
receives a reward of $1$ when visiting a new space cell, $-1$ if it hits a
wall, and $0$ otherwise.  We use a discount factor $\gamma = 0.99$, and train
the multigrid policy network using A3C~\citep{A3C}.

Figure~\ref{fig:generalization} (left) depicts the localization loss while
pre-training the mapping and localization subnets.  Freezing these subnets,
we see that localization remains reliable (Figure~\ref{fig:generalization}, middle) while reinforcement learning the policy~(Figure~\ref{fig:generalization}, right).  The results demonstrate that
the learned multigrid memory and query subnets generalize to trajectories that
differ from those in their training dataset, as also conveyed in
Table~\ref{tab:mapping} (last row).  Meanwhile, the multigrid policy network is
able to utilize memory from the mapping subnet in order to learn an effective
exploration policy.  See
\ifarxiv
   Appendix~\ref{sec:app_demo}
\else
   supplementary material
\fi
for visualizations of the exploratory behavior.

\vspace{-5pt}
\subsection{Algorithmic Tasks}

\vspace{-5pt}
We test the task-agnostic nature of our multigrid memory architecture by
evaluating on a series of algorithmic tasks, closely inspired by those
appearing in the original NTM work~\citep{NTM}.  For each of the following
tasks, we consider two variants, increasing in level of difficulty.  See
\ifarxiv
   Appendix~\ref{sec:app_exp_details_alg}
\else
   supplementary material
\fi
for complete details.

\begin{figure}[t]
    \centering
    \includegraphics[width=1.0\linewidth]{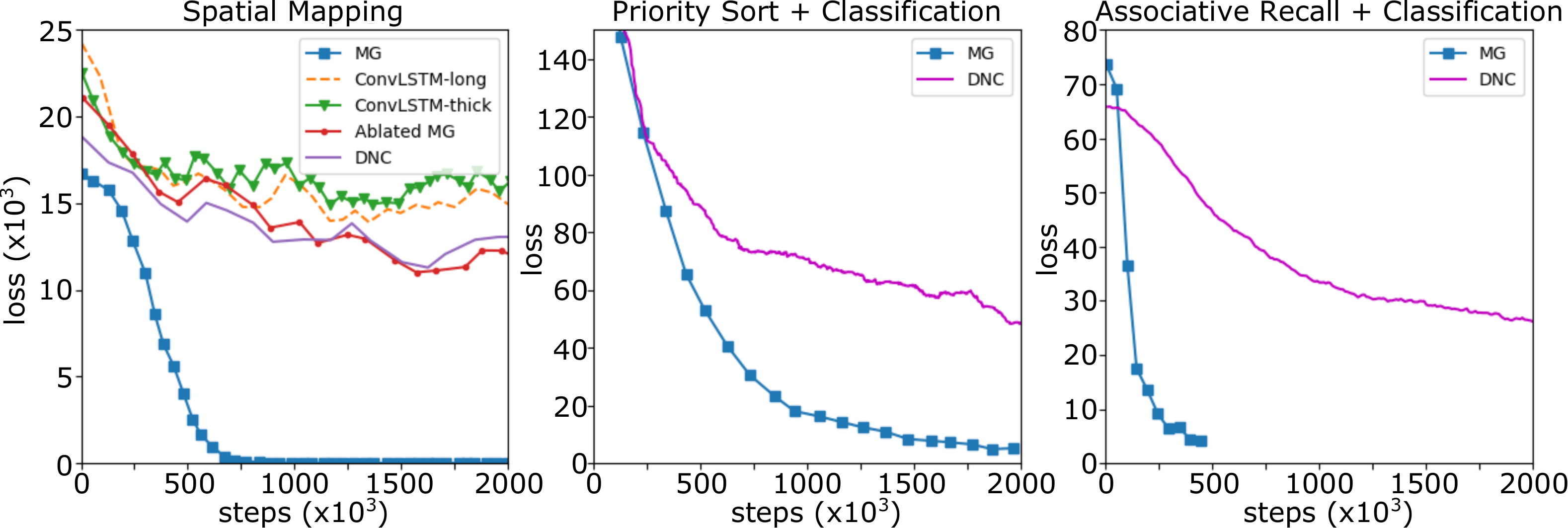}
    \vspace{-20pt}
    \caption{
        \textbf{Multigrid memory architectures learn significantly faster.}
        \emph{\textbf{Left}}:
        Maze localization task.
        \emph{\textbf{Middle}}:
        Joint priority sort and classification.
        \emph{\textbf{Right}}:
        Joint associative recall and classification.
    } \label{fig:learning_curves}
    \vspace{0pt}
\end{figure}

\begin{figure}[t]
    \centering
    \includegraphics[width=0.9\linewidth]{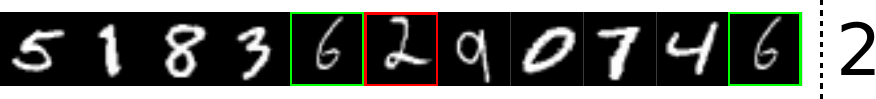}
    \vspace{-10pt}
    \caption{
     \textbf{MNIST recall.}
        A random sequence of images followed by a repeat (green),
        output the \emph{class} of the next image (red).
      }
    \vspace{-10pt}
      \label{fig:assoc_mnist}
\end{figure}

\begin{table*}[t!]
   \vspace{8pt}
   \begin{minipage}[t]{0.29\linewidth}      
      \caption{
         \textbf{Algorithmic tasks.}
         Multigrid memory architectures achieve far lower error on the
         classification variants of priority sort and associative recall,
         while performing comparably to the DNC on the simpler versions of
         these tasks. Multigrid memory architectures also remain effective when dealing with long sequences.
      } \label{tab:algs}
   \end{minipage}
   \hfill
   \begin{minipage}[t]{0.6875\linewidth}
      \vspace{0pt}
      \begin{center}
         \begin{scriptsize}
         \setlength\tabcolsep{3.5pt}
         \begin{tabular}{@{}l|c|cc|cc|c|c|c@{}}
                                  &\multirow{2}{*}{Architecture}    & \multirow{1}{*}{Params} & \multirow{1}{*}{Memory}  &           Item               & List                &  \multirow{2}{*}{Data}   & \multirow{2}{*}{Task}          & \multirow{2}{*}{Error Rate $\pm$ $\sigma$} \\
                                  &                                 &        $(\times10^6)$   &    $(\times10^3)$        &           Size               & Length              &                          &                                &                                            \\

\HLINE\tstrut
\multirow{8}{37pt}{Standard Sequence}
				  &MG Enc+Dec                       &         0.12            & \hphantom{00}7.99        & \multirow{2}{*}{1$\times$9}  & \multirow{2}{*}{20} &     Random               &  \multirow{2}{*}{Priority Sort}&  0.0043$\pm$0.0009                         \\
                                  &DNC                              &         0.76            & \hphantom{00}8.00        &                              &                     &     Patch                &                                &  0.0039$\pm$0.0016                        \\
		  
\cline{2-9}
				  &MG Enc+Dec                       &         0.29            & \hphantom{00}7.56        & \multirow{2}{*}{28$\times$28}& \multirow{2}{*}{20} &  \multirow{2}{*}{MNIST}  &  Priority Sort                 &  \textbf{0.0864$\pm$0.0016}                \\
                                  &DNC                              &         1.05            & \hphantom{00}8.00        &                              &                     &                          &  + Classify                    &  \textcolor{red}{0.1659$\pm$0.0188}        \\
			  
\cline{2-9}
                                  &MG Mem+CNN                       &         0.13            & \hphantom{00}7.99        & \multirow{2}{*}{1$\times$9}  & \multirow{2}{*}{10} &      Random              & \multirow{2}{*}{Assoc. Recall} &  0.0030$\pm$0.0002                         \\
                                  &DNC                              &         0.76            & \hphantom{00}8.00        &                              &                     &      Patch               &                                &  0.0044$\pm$0.0001                         \\
		  
\cline{2-9}
				  &MG Mem+CNN                       &         0.21            & \hphantom{00}7.56        & \multirow{2}{*}{28$\times$28}& \multirow{2}{*}{10} & \multirow{2}{*}{MNIST}   &   Assoc. Recall                &  \textbf{0.0736$\pm$0.0045}                  \\
                                  &DNC                              &         0.90            & \hphantom{00}8.00        &                              &                     &                          &   + Classify                   &  \textcolor{red}{0.2016$\pm$0.0161}        \\
\HLINE\tstrut
\multirow{2}{37pt}{Extended Sequence}
                                  &MG Enc+Dec                       &         0.89            & \hphantom{0}76.97        & \multirow{2}{*}{1$\times$9}  & 50                  &   Random                 & Priority Sort                  &  0.0067$\pm$0.0001                        \\
                                  &MG Mem+CNN                       &         0.65            & \hphantom{0}76.97        &                              & 20                  &   Patch                  & Assoc. Recall                  &  0.0056$\pm$0.0003                         \\
\hline
         \end{tabular}
         \end{scriptsize}
      \end{center}
   \end{minipage}
   \vspace{-2pt}
\end{table*}

\begin{figure*}[t]
    \centering
    \includegraphics[angle=90,origin=c,width=1.0\linewidth]{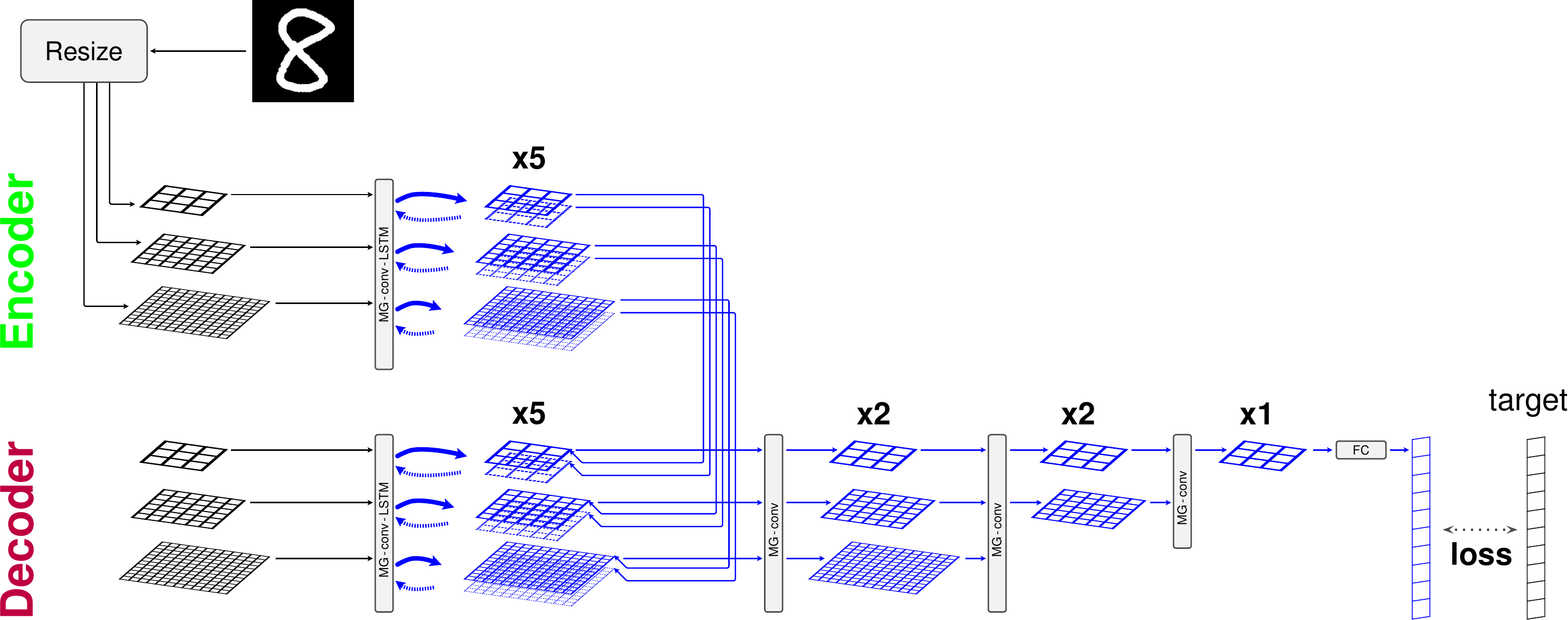}
    \vspace*{-58mm}
    \caption{
      \textbf{Multigrid memory encoder-decoder architecture for MNIST sorting.}
      After processing the input sequence, the encoder (top) transfers memory
      into the decoder, which predicts the sequence of classes of the input
      digits in sorted order.
   } \label{fig:mnist_decoder}
\end{figure*}

\begin{figure*}[h!]
    \centering
    \includegraphics[angle=90,origin=c,width=1.0\linewidth]{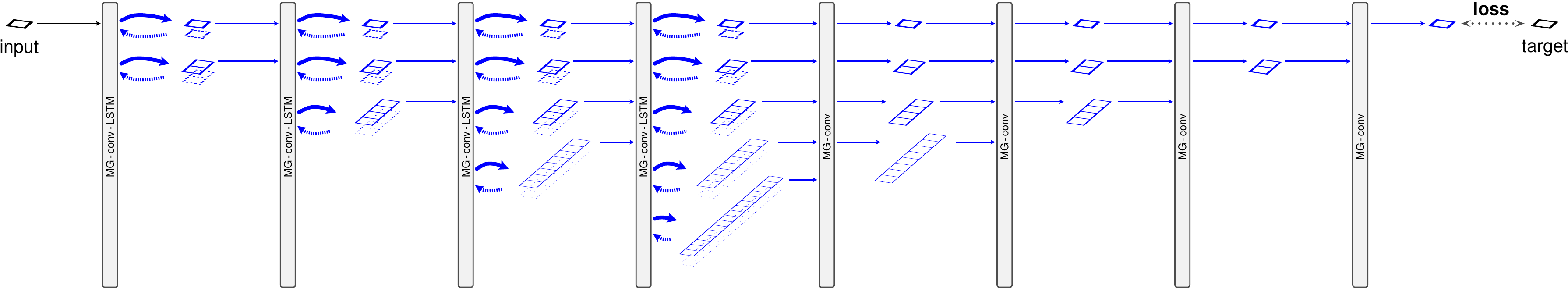}
    \vspace*{-75mm}
    \caption{
      \textbf{Multigrid memory architecture for question answering.} 1D multigrid architecture is employed for question answering tasks. Input and output are $1\times1\times159$ tensors representing the word vectors. At each time step, the model receives a word input and generates the next word in the sequence.
   } \label{fig:babi_arch}
   \vspace*{-3mm}
\end{figure*}

\begin{table*}[h!]
   \begin{minipage}[h]{0.43\linewidth}
      \caption{
         \textbf{Question answering tasks.} Despite using less memory, multigrid architecture surpasses DNC's mean performance.
         While DNC performs slightly better in the best result, its high deviation from the mean shows that DNC is much more unstable compared to multigrid memory.         
      } \label{tab:nlp}
   \end{minipage}
   \hfill
   \begin{minipage}[h]{0.55\linewidth}
      \begin{center}
         \begin{scriptsize}
         \setlength\tabcolsep{1pt}
         \begin{tabular}{@{}l|cc|c|c|c|c@{}}
                            \multirow{3}{*}{Architecture}     &                         &                          & \multicolumn{2}{c|}{Mean Result}                             &        \multicolumn{2}{c}{Best Result}     \\
\cline{4-7}                    
                     		                                   & \multirow{1}{*}{Params} & \multirow{1}{*}{Memory}  &  Mean Error $\pm$ $\sigma$ & \#Failed Tasks $\pm$ $\sigma$   &     Mean Error           & \#Failed Tasks    \\
				                                                  &        $(\times10^6)$   &    $(\times10^3)$        &     (\%)                   & (Error > 5\%)                   &        (\%)              &  (Error > 5\%)    \\

\HLINE\tstrut
				                 MG Mem                           &         1.00            & \hphantom{0}2.86         &$\hphantom{0}\bm{9.2\pm1.0}$&  $\hphantom{0}\bm{5.4\pm1.4}$   &       7.5                &        4           \\
                                  DNC                         &         0.89            & 16.38                    &           $16.7 \pm 7.6$   &        $11.2 \pm 5.4$           &       $\bm{3.8}$         &        $\bm{2}$    \\
                                  EMN                         &         N/A             & N/A                      &           N/A              &        N/A                      &       7.5                &        6           \\
\hline
         \end{tabular}
         \end{scriptsize}
      \end{center}
   \end{minipage}
   \ifarxiv
   \vspace{8pt}
   \else
   \fi
\end{table*}

\begin{table}[h!]
   \vspace{-12pt}
   \begin{minipage}[t]{1.0\linewidth}
      \vspace{0pt}
      \caption{
	      \textbf{Complete question answering results.} Shown are results for all $20$ question answering tasks.
      } \label{tab:nlp_ap}
   \end{minipage}
   \begin{minipage}[t]{1.0\linewidth}
      \vspace{3pt}
      \begin{center}
         \begin{scriptsize}
         \setlength\tabcolsep{1pt}
         \begin{tabular}{@{}l|c|c@{}}
                       		 \multirow{2}{*}{Task}           &  \multicolumn{2}{c}{$\textrm{Mean}\pm \sigma$}                     \\
\cline{2-3}
								 &  \multicolumn{1}{c|}{MG Mem}& \multicolumn{1}{c}{DNC}                \\
\HLINE\tstrut
				  single supporting fact         &  $\hphantom{0}\bf{0.3 \pm 0.3}\hphantom{0}$    & $\hphantom{0}9.0 \pm 12.6$              \\
				  two supporting facts           &  $\bf{27.4 \pm 7.8}\hphantom{0}$               & $39.2 \pm 20.5$             \\
				  three supporting facts         &  $51.8 \pm 5.0\hphantom{0}$                    & $\bf{39.6 \pm 16.4}$             \\
				  two argument relations         &  $\hphantom{0}\bf{0.1 \pm 0.1}\hphantom{0}$    & $\hphantom{0}0.4 \pm 0.7\hphantom{0}$               \\
				  three argument relations       &  $\hphantom{0}3.4 \pm 0.6\hphantom{0}$         & $\hphantom{0}\bf{1.5 \pm 1.0}\hphantom{0}$                \\
				  yes/no questions               &  $\hphantom{0}\bf{1.7 \pm 1.0}\hphantom{0}$    & $\hphantom{0}6.9 \pm 7.5\hphantom{0}$                \\
				  counting                       &  $\hphantom{0}\bf{5.2 \pm 2.1}\hphantom{0}$    & $\hphantom{0}9.8 \pm 7.0\hphantom{0}$                \\
				  lists/sets                     &  $\hphantom{0}\bf{4.5 \pm 1.3}\hphantom{0}$    & $\hphantom{0}5.5 \pm 5.9\hphantom{0}$                \\
				  simple negation                &  $\hphantom{0}\bf{0.8 \pm 0.6}\hphantom{0}$    & $\hphantom{0}7.7 \pm 8.3\hphantom{0}$                \\
				  indefinite knowledge           &  $\hphantom{0}\bf{5.9 \pm 1.9}\hphantom{0}$    & $\hphantom{0}9.6 \pm 11.4$               \\
				  basic coreference              &  $\hphantom{0}\bf{0.5 \pm 0.2}\hphantom{0}$    & $\hphantom{0}3.3 \pm 5.7\hphantom{0}$                \\
				  conjunction                    &  $\hphantom{0}\bf{0.4 \pm 0.3}\hphantom{0}$    & $\hphantom{0}5.0 \pm 6.3\hphantom{0}$                \\
				  compound coreference           &  $\hphantom{0}\bf{0.1 \pm 0.1}\hphantom{0}$    & $\hphantom{0}3.1 \pm 3.6\hphantom{0}$                \\
				  time reasoning                 &  $22.0 \pm 1.9\hphantom{0}$                    & $\bf{11.0 \pm 7.5}\hphantom{0}$               \\
				  basic deduction                &  $\bf{0.4 \pm 0.5}$                            & $27.2 \pm 20.1$              \\
				  basic induction	             &  $\bf{51.6 \pm 2.9}\hphantom{0}$               & $53.6 \pm 1.9\hphantom{0}$              \\
				  positional reasoning 	         &  $\hphantom{0}\bf{4.6 \pm 2.1}\hphantom{0}$    & $32.4 \pm 8.0\hphantom{0}$               \\
				  size reasoning	             &  $\hphantom{0}\bf{1.7 \pm 0.7}\hphantom{0}$    & $\hphantom{0}4.2 \pm 1.8\hphantom{0}$                \\
				  path finding	                 &  $\bf{1.4 \pm 1.2}$                            & $64.6 \pm 37.4$              \\
		 		  agents motivations	         &  $\hphantom{0}0.2 \pm 0.4\hphantom{0}$         & $\hphantom{0}\bf{0.0 \pm 0.1}\hphantom{0}$                \\
\HLINE\tstrut
				  Mean Error (\%)                &  $\hphantom{0}\bf{9.2 \pm 1.0}\hphantom{0}$    & $16.7 \pm 7.6\hphantom{0}$            \\
				  Failed Tasks (Error > 5\%)     &  $\hphantom{0}\bf{5.4 \pm 1.4}\hphantom{0}$    & $11.2 \pm 5.4\hphantom{0}$            \\

\hline
         \end{tabular}
         \end{scriptsize}
      \end{center}
   \end{minipage}
   \vspace{-10pt}
\end{table}

\textbf{Priority Sort.}
In the first non-visual variant, the network receives a sequence of twenty
$9$-dimensional vectors, along with their priority.  The task is to output the
sequence of vectors in order of their priority.  Training and testing use
randomly generated data.  Training takes $2\times10^6$ steps, batch size $32$,
and testing uses $5000$ sequences.  Results are computed over $10$ runs.  We
tune hyperparameters as done for the mapping task.  We structure our model as
an encoder-decoder architecture (Figure~\ref{fig:arch_nets}, bottom).  Our
network performs equivalently to DNC with equal memory, with both achieving
near-perfect performance (Table~\ref{tab:algs}).

The second variant extends the priority sort to require recognition capability.
The input is a sequence of twenty $28 \times 28$ MNIST images~\citep{MNIST}.
The goal is to output the \emph{class} of the input images in increasing order.
A sample encoder-decoder architecture employed for this task is presented in Figure~\ref{fig:mnist_decoder}.
Table~\ref{tab:algs} reveals that our architecture achieves much lower error
rate compared to DNC on this task (priority sort + classification), while also
learning faster (Figure~\ref{fig:learning_curves}) and with less memory.

\textbf{Associative Recall.}
In the first formulation, the network receives a sequence of ten $9$-element
random vectors, followed by a second instance of one of the first nine vectors.
The task is to output the vector that immediately followed the query in the
input sequence.  We demonstrate this capability using the multigrid
reader/writer architecture (Figure~\ref{fig:arch_nets}, top).  Training is
similar to the sorting task.  Table~\ref{tab:algs} shows that both DNC and our
architecture achieve near-zero error rates.

In the second variant, the input is a sequence of ten \mbox{$28 \times 28$} randomly
chosen MNIST images~\citep{MNIST}, where the network needs to output the
\emph{class} of the image immediately following the query (Figure~\ref{fig:assoc_mnist}).  As shown in
Table~\ref{tab:algs} and Figure~\ref{fig:learning_curves}, our multigrid
memory network performs this task with significantly greater accuracy than the
DNC, and also learns in fewer training steps.

To further test the ability of multigrid memory architectures to deal with
longer sequences, we experimented with sorting and associative recall with
sequence length of $50$ and $20$, respectively.  As can be seen in
Table~\ref{tab:algs}, multigrid memory architectures remain effective with
near-zero error rates.

The harder variants of both priority sort and associative recall require a
combination of memory and a pattern recognition capability.  The success of
multigrid memory networks (and notable poor performance of DNCs), demonstrates
that they are a unique architectural innovation.  They are capable of learning
to simultaneously perform representational transformations and utilize a large
distributed memory store.  Furthermore, as Figure~\ref{fig:learning_curves}
shows, across all difficult tasks, including mapping and localization,
multigrid memory networks train substantially faster and achieve substantially
lower loss than all competing methods.

\subsection{Question Answering}
\label{sec:nlp-evaluation}

To further investigate the generalizability of our multigrid memory
architecture well beyond spatial reasoning, we evaluate its performance on
bAbI~\cite{weston15}, which consist of $20$ question answering tasks
corresponding to diverse aspects of natural language understanding. For each task, the
dataset contains $10000$ questions for training and $1000$ for testing. For parameter-efficiency, we employ a 1D multigrid architecture for this task, as shown in Figure~\ref{fig:babi_arch}. Results are shown in Table~\ref{tab:nlp}.  Despite
having only a fraction of the memory available to the DNC
($2.86$K v.s.\ $16.38$K), on average our architecture outperforms the DNC in terms of
both mean error rate ($9.2\%$ v.s.\ $16.7\%$) and mean failed tasks
($5.4$ v.s.\ $11.2$). On best results, while DNC performs slightly better, multigrid memory still outperforms EMN~\citep{sukhbaatar2015end} with fewer failed tasks. Complete results for all tasks are presented in Table~\ref{tab:nlp_ap}. These results not only demonstrate the adaptability of
multigrid memory, but are also a testament to our design's effectiveness.  See
\ifarxiv
   Appendix~\ref{sec:app_nlp}
\else
   supplementary material
\fi
for further details.

\section{Conclusion}
\label{sec:conclusion}

\vspace{-5pt}
Multigrid memory represents a radical new approach to augmenting networks
with long-term, large-scale storage.  A simple principle, multigrid connectivity, underlies
our model.  Residual networks~\cite{he2015deep}, which provide shortcut
pathways across depth, have had enormous impact, allowing very deep networks to
be trained by facilitating gradient propagation.  Multigrid wiring is
complementary, improving connectivity across an orthogonal aspect of the
network: the spatial dimension.  Its impact is equally significant: multigrid
wiring exponentially improves internal data routing efficiency, allowing
complex behaviors (attention) and coherent memory subsystems to emerge from
training simple components.  Our results are cause to rethink the
prevailing design principles for neural network architectures.

\ifdefined\isaccepted
   \textbf{Acknowledgments.}
We thank Gordon Kindlmann for his support in pursuing this project,
Chau Huynh for her help with the code, and Pedro Savarese and Hai Nguyen for
fruitful discussions.  The University of Chicago CERES Center contributed to
the support of Tri Huynh.  This work was supported in part by the National
Science Foundation under grant IIS-1830660.

\else
   \ifarxiv
   \textbf{Acknowledgments.}
We thank Gordon Kindlmann for his support in pursuing this project,
Chau Huynh for her help with the code, and Pedro Savarese and Hai Nguyen for
fruitful discussions.  The University of Chicago CERES Center contributed to
the support of Tri Huynh.  This work was supported in part by the National
Science Foundation under grant IIS-1830660.

   \fi
\fi

{\small
\bibliographystyle{icml2020}
\ifarxiv
   \bibliography{nmg-mem_arxiv}
\else
   \bibliography{nmg-mem}
\fi
}
\flushcolsend

\ifarxiv
    \clearpage
    \appendix
\section{Information Routing} \label{sec:appendix}
\label{sec:app_information_routing}

\begin{figure*}[t]
   \includegraphics[width=1.0\linewidth]{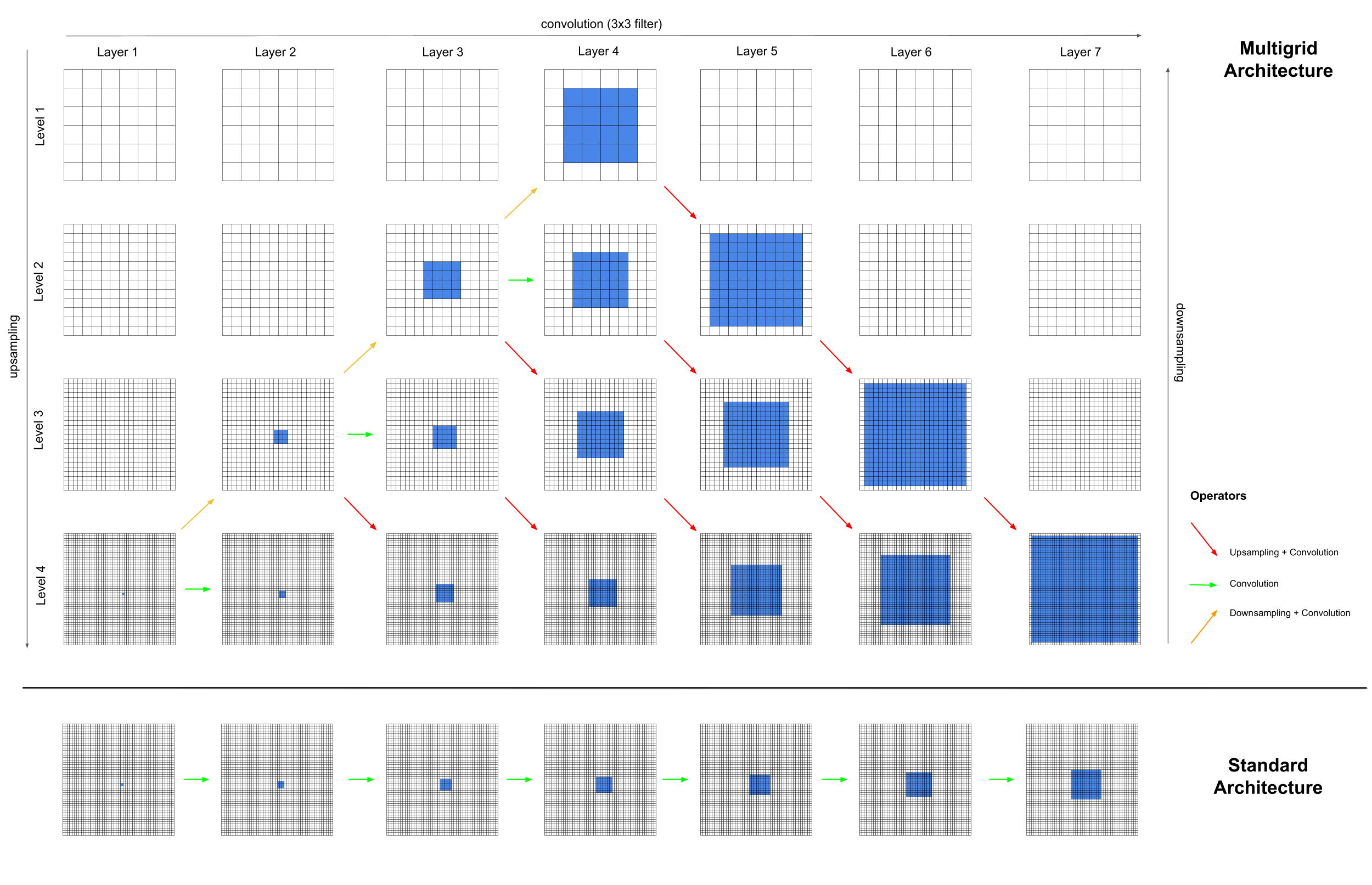}\\
   \vspace{-25pt}
   \caption{
      \textbf{Information routing.}
      \emph{\textbf{Top:}}
      Paths depicting information flow in a multigrid architecture.
      Progressing from one layer to the next, information flows between grids
      at the same level (via convolution, \textcolor{green}{green}), as well as
      to adjacent grids at higher resolution (via upsampling and convolution,
      \textcolor{red}{red}) and lower resolution (via downsampling and
      convolution, \textcolor{orange}{orange}).  Information from a sample
      location $(26,26)$ (\textcolor{blue}{blue}) of the source grid at
      [layer 1, level 4] can be propagated to all locations rendered in
      \textcolor{blue}{blue} in subsequent layers and levels, following the
      indicated paths (among others).
      Information quickly flows from finer levels to coarser levels, and then
      to any location in just a few layers.  Receptive field size grows
      exponentially with depth.  In practice, the routing strategy is
      emergent---routing is determined by the learned network parameters
      (convolutional filters).  Multigrid connectivity endows the network with
      the potential to quickly route from any spatial location to any other
      location just a few layers deeper.
      \emph{\textbf{Bottom:}}
      Information flow in a standard architecture.  Without multigrid
      connections, information from the same source location is propagated
      much more slowly across network layers.  Receptive fields expand at a
      constant rate with depth, compared to the multigrid network's exponential
      growth.
   }
   \label{fig:information_routing}
\end{figure*}

\emph{\textbf{Proposition 1:}}
For the setup in Figure~\ref{fig:information_routing}, suppose that the
convolutional kernel size is $3\times3$, and upsampling is $2\times$
nearest-neighbor sampling. Consider location $(1,1)$ of the source grid at
\mbox{[layer $1$, level $1$]}.
For a target grid at [layer $m$, level $n$],
where $m \geq n$, the information from the source location can be routed to any
location $(i,j)$, where $1 \leq  i$, $j  \leq   (m-n+2) \cdot 2^{n-1} - 1$.

\emph{\textbf{Proof of Proposition 1:}}
Induction proof on level $n$.
\begin{itemize}[leftmargin=0.2in]
   \item{%
      For level $n = 1$: Each convolution of size $3\times3$ can direct
      information from a location $(i,j)$ at layer $k$ to any of its immediate
      neighbors $(i',j')$ where
         $i-1 \leq  i' \leq   i+1$, $j-1 \leq  j' \leq j+1$ in layer $k+1$.
      Therefore, convolutional operations can direct information from location
      $(1,1)$ in layer $1$ to any locations $(i',j')$ in layer $k=m$ where
         $1 \leq  i'$ , $j' \leq m = (m-1+2) \cdot 2^0 - 1
                                   = (m-n+2) \cdot 2^{n-1} - 1$.
   }
   \item{
      Assume the proposition is true for level $n$ ($\forall m \geq n$), we
      show that it is true for level $n+1$.  Consider any layer $m+1$ in level
      $n+1$, where $m+1 \geq n+1$:

      We have, $m+1 \geq n+1 \Rightarrow m \geq n$.  Therefore, we have that at
      [layer $m$, level $n$],
      the information from the source location can be routed to any location
      $(i,j)$, where $1 \leq  i$, $j \leq (m-n+2) \cdot 2^{n-1} - 1$.  Now,
      consider the path from
      [layer $m$, level $n$]
      to
      [layer $m+1$, level $n+1$].
      This path involves the upsampling followed by a convolution operator, as
      illustrated in Figure~\ref{fig:information_routing}.

      Nearest-neighbor upsampling directly transfers information from index
      $i$ to $2 \cdot i$ and $2 \cdot i-1$, and
      $j$ to $2 \cdot j$ and $2 \cdot j-1$ by definition.
      For simplicity, first consider index $i$ separately.
      By transferring to $2 \cdot i$, information from location
      $1 \leq  i \leq  (m-n+2) \cdot 2^{n-1} - 1$ in level $n$ will be
      transferred to all even indices in
      $[2, ((m-n+2) \cdot 2^{n-1} - 1)\cdot 2]$ at level $n+1$.
      By transferring to $2 \cdot i - 1$, information from location
      $1 \leq  i \leq  (m-n+2) \cdot 2^{n-1} - 1$ in level $n$ will be
      transferred to all odd indices in
      $[1, ((m-n+2) \cdot 2^{n-1} - 1)\cdot 2 - 1]$ at level $n+1$.
      Together, with $2 \cdot i$ and $2 \cdot i-1$ transferring, the
      nearest-neighbor upsampling transfers information from location
      $1 \leq i \leq (m-n+2) \cdot 2^{n-1} - 1$ in level $n$ to all indices in
      $[1, ((m-n+2) \cdot 2^{n-1} - 1)\cdot 2]$ at level $n+1$.

      Furthermore, the following convolution operator with $3\times3$ kernel
      size can continue to transfer information from
      $[1, ((m-n+2) \cdot 2^{n-1} - 1)\cdot 2]$ to
      $[1, ((m-n+2) \cdot 2^{n-1} - 1)\cdot 2+1]$ at level $n+1$.
      We have
      $((m-n+2) \cdot 2^{n-1} - 1)\cdot 2+1 = (m+1-(n+1)+2) \cdot 2^n - 1$.
      Taking together indices $i$ and $j$, information from location $(i,j)$
      where $1 \leq  i$, $j  \leq   (m-n+2) \cdot 2^{n-1} - 1$ in level $n$
      can be transferred to $(i’,j’)$ in level $n+1$, where
      $1 \leq  i'$, $j' \leq (m+1-(n+1)+2) \cdot 2^n - 1$. $\blacksquare$
   }
\end{itemize}

\section{Experiment Details}
\label{sec:app_exp_details}

\begin{figure*}[t]
   \centering
   \includegraphics[angle=0,origin=c,width=1.0\linewidth]{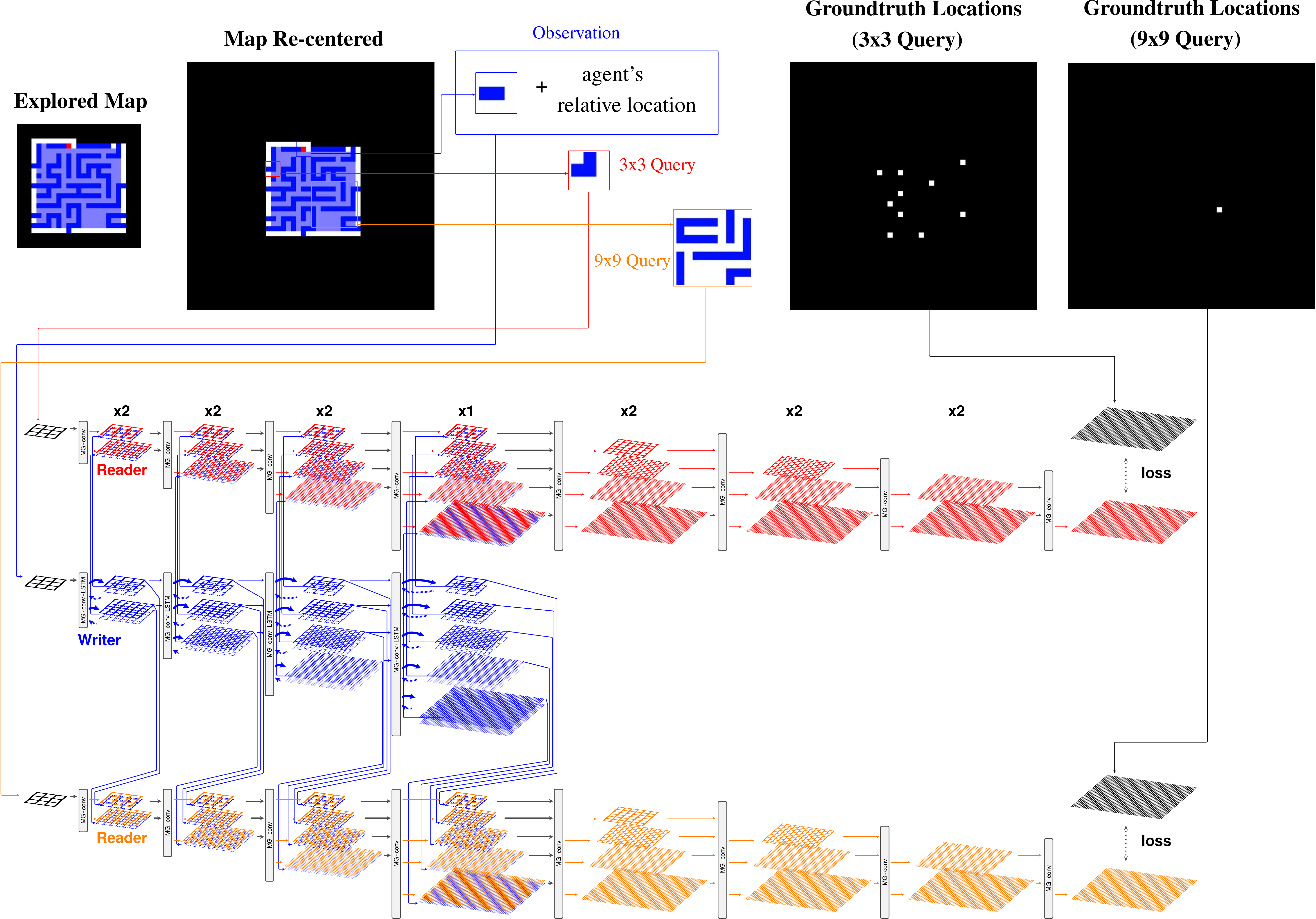}
   \vspace*{0mm}
   \caption{
      \textbf{Multigrid memory writer-reader(s) architecture for spatial
      navigation.}
      At each time step, the agent moves to a new location and observes the
      surrounding $3\times3$ patch.  The writer receives this $3\times3$
      observation along with the agent's relative location (with respect to
      the starting point), updating the memory with this information.  Two
      readers receive randomly chosen $3\times3$ and $9\times9$ queries, view
      the current map memory built by the writer, and infer the possible
      locations of those queries.
   }
   \label{fig:rw_mapping}
\end{figure*}

\subsection{Spatial Navigation}
\label{sec:app_exp_details_sn}
\subsubsection{Architecture}

All experiments related to spatial navigation tasks use multigrid
writer-reader(s) architectures.  Figure~\ref{fig:rw_mapping} visualizes this
architecture and problem setup.  At each time step during training, the agent
takes a one-step action (\eg~along a spiral trajectory) and observes its
$3\times3$ surroundings.  This observation, together with its location
relative to the starting point, are fed into the writer, which must learn to
update its memory.  The agent has no knowledge of its absolute location in the
world map.  Two random $3\times3$ and $9\times9$ patches within the explored
map are presented to the agent as queries (some experiments use only
$3\times3$ queries).  These queries feed into two readers, each viewing the
same memory built by the writer; they must infer which previously seen
locations match the query.  Since the agent has no knowledge of its absolute
location in the world map, the agent builds a map relative to its initial
position (\textit{map re-centered} in Figure~\ref{fig:rw_mapping}) as it
navigates.

During training, the writer learns to organize and update memory from
localization losses simultaneously backpropagated from the two readers.  During
inference, only the writer updates the memory at each time step, and the
readers simply view (\ie~without modification) the memory to infer the query
locations.  It is also worth noting that only $3\times3$ patches are fed into
the writer at each time step; the agent never observes a $9\times9$ patch.
However, the agent successfully integrates information from the $3\times3$
patches into a coherent map memory in order to correctly answer queries much
larger than its observations.
\ifarxiv
    Figure~\ref{fig:learning_rep}
\else
    Figure 4 in the main document
\fi
shows that this
learned memory strikingly resembles the actual world map.

\subsubsection{Loss}
\label{sec:app_spatial_map_loss}

Given predicted probabilities and the ground-truth location mask
(Figure~\ref{fig:rw_mapping}), we employ a pixel-wise cross-entropy loss as the
localization loss.  Specifically, letting $S$ be the set of pixels, $p_i$ be
the predicted probability at pixel $i$, and $y_i$ be the binary ground-truth at
pixel $i$, the pixel-wise cross-entropy loss is computed as follows:
\begin{equation}
-\sum_{i\in S} y_i \log(p_i)+(1-y_i)\log(1-p_i)
\end{equation}

\subsection{Algorithmic Tasks}
\label{sec:app_exp_details_alg}
\subsubsection{Architecture}

\paragraph{Priority Sort Tasks:}
We employ encoder-decoder architectures for the priority sort tasks.

\begin{itemize}
   \item{
      \textit{Standard variant}.
      The encoder is a 5-layer multigrid memory architecture, structured similar to the writer in Figure~\ref{fig:rw_mapping}, progressively scaling $3\times3$ inputs at the coarsest resolution into $24\times24$ resolution.  For the decoder, the first half of the
      layers (MG-conv-LSTM) resemble the encoder, while the second half employ
      MG-conv layers to progressively scale down the output to $3\times3$.
   }
   \item{
      \textit{MNIST sort + classification}.
      \ifarxiv
         Figure~\ref{fig:mnist_decoder}
      \else
         Figure 8 in the main document
      \fi depicts the encoder-decoder architecture
      for the MNIST variant.
   }
\end{itemize}

\paragraph{Associative Recall Tasks:}
We employ writer-reader architectures for the associative recall tasks.  The
architectures are similar to those for the spatial navigation and priority sort tasks depicted in
Figure~\ref{fig:rw_mapping}, with some modifications appropriate to the tasks:
\begin{itemize}
   \item{
      \textit{Standard variant}.
      In the standard version of the task, the writer architecture is similar to the encoder in the standard variant of the priority sort task.  For the reader, after progressing
      to the finest resolution corresponding to the memory in
      the writer, the second half of MG-conv layers progressively scale down
      the output to $3\times3$ to match the expected output size.
   }
   \item{
      \textit{MNIST recall + classification}.
      For the MNIST variant, we resize the $28\times28$ images to three scales from $3\times3$ to $12\times12$ and maintain the
      same three-scale structure for five layers of the writer.  The writer
      architecture is similar to the encoder architecture in MNIST priority
      sort task, as depicted in
      \ifarxiv
         Figure~\ref{fig:mnist_decoder}.
      \else
         Figure 8 in the main document.
      \fi The reader for
      the MNIST variant is similar to the reader in the standard variant, with
      the final layer followed by a fully connected layer to produce a
      10-way prediction vector over MNIST classes.
   }
\end{itemize}

\subsubsection{Loss}
\label{sec:app_alg_loss}

\paragraph{Standard variants:}
We use pixel-wise cross-entropy loss for the standard variants, as described in
Section~\ref{sec:app_spatial_map_loss}.

\paragraph{MNIST variants:}
For MNIST variants, we use cross-entropy loss over a softmax prediction of the
classes.  Specifically, letting $C$ be the set of available classes, $p_c$ the
softmax output for class $c$, and $y$ a one-hot vector of the ground-truth
label, we compute the loss as:
\begin{equation}
-\sum_{c\in C} y_c \log(p_c)
\end{equation}

\subsection{Question Answering}
\label{sec:app_nlp}
\subsubsection{Architecture}
We employ a 1D multigrid memory architecture for question answering tasks, where the spatial dimensions progressively scale from $1\times1$ to $1\times16$ through MG-conv-LSTM layers, and gradually scale back to $1\times1$ through MG-conv layers, as demonstrated in
      \ifarxiv
         Figure~\ref{fig:babi_arch}.
      \else
         Figure 9 in the main document.
      \fi
Inputs and outputs are $1 \times 1 \times \lvert V \rvert$ tensors representing the word vectors, where $V$ is the set of words in the vocabulary and $\lvert V \rvert =159$. All 20 question answering tasks are jointly trained, with batch size 1, and sequence-wise normalization. At each time step, the model receives a word input and generates the next word in the sequence. During training, only the losses from words corresponding to answers are backpropagated, others are masked out, as specified next.

\subsubsection{Loss}
Let $V$ be the set of words in the vocabulary, and $y\in \{0,1\}^{\lvert V \rvert}$ be a one-hot vector that represents the ground-truth word. For a word sequence $S$, we define a mask $m$ as:
\begin{equation}
    m_i=
    \begin{cases}
      1 & \text{if word $i$ in the sequence $S$ is an answer} \\
      0 & \text{otherwise}
    \end{cases}
\end{equation}

Letting $p\in (0,1)^{\lvert V \rvert}$ be the softmax output, we compute the loss for question answering as follows:

\begin{equation}
-\sum_{i=1}^{\lvert S \rvert}m_i\sum_{j=1}^{\lvert V \rvert}y_j^i\log(p_j^i)
\end{equation}


\section{DNC Details}
\label{sec:app_dnc}

\begin{figure}[t]
   \vspace{0pt}
   \hfill
   \begin{minipage}[t]{1.0\linewidth}
   \begin{center}
      \vspace{0pt}
      \includegraphics[width=0.8\linewidth]{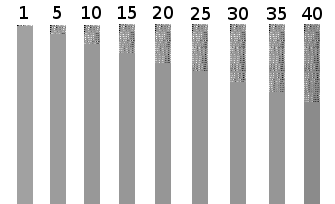}
   \end{center}
   \end{minipage}
   \vspace{-3pt}
   \begin{minipage}[t]{1.0\linewidth}
   \vspace{0pt}
   \caption{
      \textbf{Visualization of DNC memory in mapping task.}
      Due to its defined addressing mechanism, the DNC always allocates a
      new continuous memory slot at each time-step.  It does not appear to
      maintain an interpretable structure of the map.
   }
   \label{fig:dnc_mem}
\end{minipage}
\end{figure}

We use the official DNC implementation (\url{https://github.com/deepmind/dnc}),
with $5$ controller heads ($4$ read heads and $1$ write head). For spatial navigation and algorithmic tasks,
we use a memory vector of $16$ elements, and $500$ memory slots ($8$K total), which is the largest memory
size permitted by GPU resource limitations.  Controllers are LSTMs, with hidden
state sizes chosen to make total parameters comparable to other models in
\ifarxiv
Table~\ref{tab:mapping} and Table~\ref{tab:algs}.
\else
Table 1 and Table 2 in the main document.
\fi
DNC imposes a relatively small cap on the addressable memory due to the quadratic cost of the temporal
linkage matrix
(\url{https://github.com/deepmind/dnc/blob/master/dnc/addressing.py#L163}).

A visualization of DNC memory in the spatial mapping task ($15\times15$ map) is
provided in Figure~\ref{fig:dnc_mem}.

For question answering tasks, the DNC memory is comprised of $256$ memory slots, with a $64$-element vector for each slot ($16,384$ total). The use of a smaller number of memory slots and batch size allows for the allocation of larger total memory.

\section{Runtime}
\label{sec:app_runtime}

On spatial mapping (with $15\times15$ world map), the runtimes for one-step
inference with the Multigrid Memory architecture ($0.12$\,M parameters and
$8$\,K memory) and DNC ($0.75$\,M parameters and $8$\,K memory) are
($\text{mean} \pm \text{std}$): $0.018 \pm 0.003$ seconds and $0.017 \pm 0.001$ seconds, respectively.
These statistics are computed over $10$ runs on a NVIDIA Geforce GTX Titan X.


\section{Demos}
\label{sec:app_demo}

\begin{itemize}[leftmargin=0.2in]
   \item Instructions for interpreting the video demos:\\
   {\small{\url{https://drive.google.com/file/d/18gvQRhNaEbdiV8oNKOsuUXpF75FEHmgG}}}
   \item Mapping \& localization in spiral trajectory, with $3\times3$ queries:\\
   {\small{\url{https://drive.google.com/file/d/1VGPGHqcNXBRdopMx11_wy9XoJS7REXbd}}}
   \item Mapping \& localization in spiral trajectory, with $3\times3$ and $9\times9$ queries:\\
   {\small{\url{https://drive.google.com/file/d/18lEba0AzpLdAqHhe13Ah3fL2b4YEyAmF}}}
   \item Mapping \& localization in random trajectory:\\
   {\small{\url{https://drive.google.com/file/d/19IX93ppGeQ56CqpgvN5MJ2pCl46FjgkO}}}
   \item Joint exploration, mapping \& localization:\\
   {\small{\url{https://drive.google.com/file/d/1UdTmxUedRfC-E6b-Kz-1ZqDRnzXV4PMM}}}\\
\end{itemize}

\fi

\end{document}